\documentclass[pdflatex,sn-mathphys-num]{sn-jnl}% Math and Physical Sciences Numbered Reference Style

\usepackage{graphicx}%
\usepackage{multirow}%
\usepackage{amsmath,amssymb,amsfonts}%
\usepackage{amsthm}%
\usepackage{mathrsfs}%
\usepackage[title]{appendix}%
\usepackage{xcolor}%
\usepackage{textcomp}%
\usepackage{manyfoot}%
\usepackage{booktabs}%
\usepackage{algpseudocode}%
\usepackage{listings}%
\usepackage{svg}
\usepackage{caption,adjustbox}
\usepackage[utf8]{inputenc}
\usepackage{graphicx,color}
\usepackage{amsmath}
\usepackage[version=4]{mhchem}
\usepackage{siunitx}
\usepackage{longtable,tabularx}
\usepackage{subcaption}
\usepackage{hyperref} 
\usepackage{arydshln}
\usepackage{svg}
\usepackage{optidef} % to write optimization problems
\usepackage[ruled,vlined]{algorithm2e}  % to write algorithms
\usepackage{mathrsfs}   % to write cursive letters
\usepackage{changepage}   % to prevent overfull box
\usepackage{float} 
\usepackage{dsfont}
\usepackage{tabularx}
\usepackage{booktabs}
\usepackage{longtable}
\usepackage{adjustbox} % Potential solution?
\usepackage{amsmath}     

\usepackage{amsthm}
\usepackage{csquotes}
\MakeOuterQuote{"}

\usepackage{booktabs}

\usepackage{subcaption}

\newcommand{\eqnref}[1]{Eq.~(\ref{#1})}
\usepackage{bm}

\usepackage{optidef} % to write optimization problems
\usepackage[ruled,vlined]{algorithm2e}  % to write algorithms
\usepackage{mathrsfs}   % to write cursive letters
\usepackage{changepage}   % to prevent overfull box
\usepackage{float} 
\usepackage{dsfont}

\usepackage{amsthm}
\usepackage{comment}

\newcommand{\x}{{\bm{x}}}

\newcommand{\f}{{\bm{f}}}
\newcommand{\hyi}{\hat{y}_i(\x)}
\newcommand{\hsi}{s_i^2(\x)}

\theoremstyle{thmstyleone}%

\theoremstyle{thmstyletwo}%

\theoremstyle{thmstylethree}%
\usepackage{caption}   % for captionof if you prefer
\newenvironment{tableplain}[1][]{%
  \begin{tableorg}[#1]%
    \begin{center}%
      \tablebodyfont%
      \renewcommand\footnotetext[2][]{{\removelastskip\vskip3pt%
        \let\tablebodyfont\tablefootnotefont%
        \hskip0pt\if!##1!\else{\smash{$^{##1}$}}\fi##2\par}}%
}{%
    \end{center}%
  \end{tableorg}%
}

\begin{document}

\title[Hierarchical Bayesian optimization of an aircraft-based multi-agent 
system-of-systems]
{Hierarchical Bayesian optimization of an aircraft-based multi-agent 
system-of-systems}

\author*[1]{Paul Saves}\email{paul.saves@irit.fr}

\author[2,3]{Thierry Lefebvre}\email{ thierry.lefebvre@onera.fr}
\author[2,3]{Nathalie Bartoli}\email{nathalie.bartoli@onera.fr}

\author[4]{\\ Jasper H. Bussemaker }\email{jasper.bussemaker@dlr.de}
\author[4]{Nikolaos Kalliatakis}\email{Nikolaos.Kalliatakis@dlr.de}
\author[4]{Nabih Naeem}\email{Nabih.Naeem@dlr.de}
\author[4]{Prajwal Prakasha}\email{Prajwal.Prakasha@dlr.de}

\affil[1]{IRIT, UMR 5505 CNRS, Université Toulouse Capitole, Université de Toulouse, Toulouse, France}
\affil[2]{ONERA, DTIS, Universit\'e de Toulouse, Toulouse, France}
\affil[3]{F\'ed\'eration ENAC ISAE-SUPAERO ONERA, Universit\'e de Toulouse, 31000, Toulouse, France}
\affil[4]{German Aerospace Center (DLR), Institute of System Architectures in Aeronautics.
Hein-Saß-Weg 22, 21129 Hamburg, Germany}

\abstract{
Developing innovative system architectures increasingly relies on advanced modeling and optimization techniques to frame the architecting process and define the corresponding computational problems. 
In the context of complex System-of-Systems (SoS), high-fidelity multiphysics and multidisciplinary simulations are essential for capturing detailed behaviors. However, their severe computational expense and the risk of evaluation failures make direct optimization highly challenging. To overcome these limitations, surrogate-based approaches, particularly Bayesian optimization, have emerged as highly effective tools for managing expensive, black-box simulation tasks.
This work introduces a hierarchical Bayesian optimization framework that leverages Gaussian process meta-modeling to handle discrete architectural choices, conditional dependencies, and heterogeneous design variables inherent to SoS problems. Results show that the hierarchical formulation improves search efficiency and robustness compared to conventional surrogate-based methods, enabling the exploration of large and structurally diverse design spaces with limited simulation budgets.

The approach is demonstrated through the optimization of an aircraft-based multi-agent system for wildfire suppression, a use case developed within the EU-funded COLOSSUS project that illustrates how SoS principles can be applied to coordinate heterogeneous aerial platforms with complementary roles, supporting both sustainable mobility and emergency response missions.
Our framework provides a scalable methodology for SoS architecting and model exploration, offering transferable insights for applications in aviation, sustainable mobility, and resilience-oriented system design. By combining hierarchical representations with surrogate-based optimization, this work is among the first practical demonstrations of hierarchical Bayesian optimization applied to real-world SoS problems, advancing both methodology and practice.

}


\maketitle

\section*{Nomenclature}
\begin{center}
\begin{tabular}{ll}
%\hline
ABM & Agent-Based Models \\
ABS & Agent-Based Simulations \\
ADAM & Advanced Air Mobility \\
ADSG & Architecture Design Space Graph \\
BO & Bayesian Optimization \\
ConOps&Concepts of Operations \\
DoE & Design of Experiments \\
EIS & Entry Into Service \\
EVE& Eco-friendly Vehicle for multiple operating Environments \\
eVTOL & Electric Vertical Take-Off and Landing \\
GP & Gaussian Process\\
KPI & Key Performance Indicator\\
PF & Pareto Front \\
SMT & Surrogate Modeling Toolbox \\
SoS & System-of-Systems \\
SoSID &System-of-Systems Inverse Design \\
TLAR & Top-Level Aircraft Requirements \\ 
%\hline
\end{tabular}
\end{center}

\section{Introduction}
\renewcommand*\footnoterule{}
\label{sec:intro}

\begin{figure}[H]
\begin{center}
\includegraphics[scale=0.55]{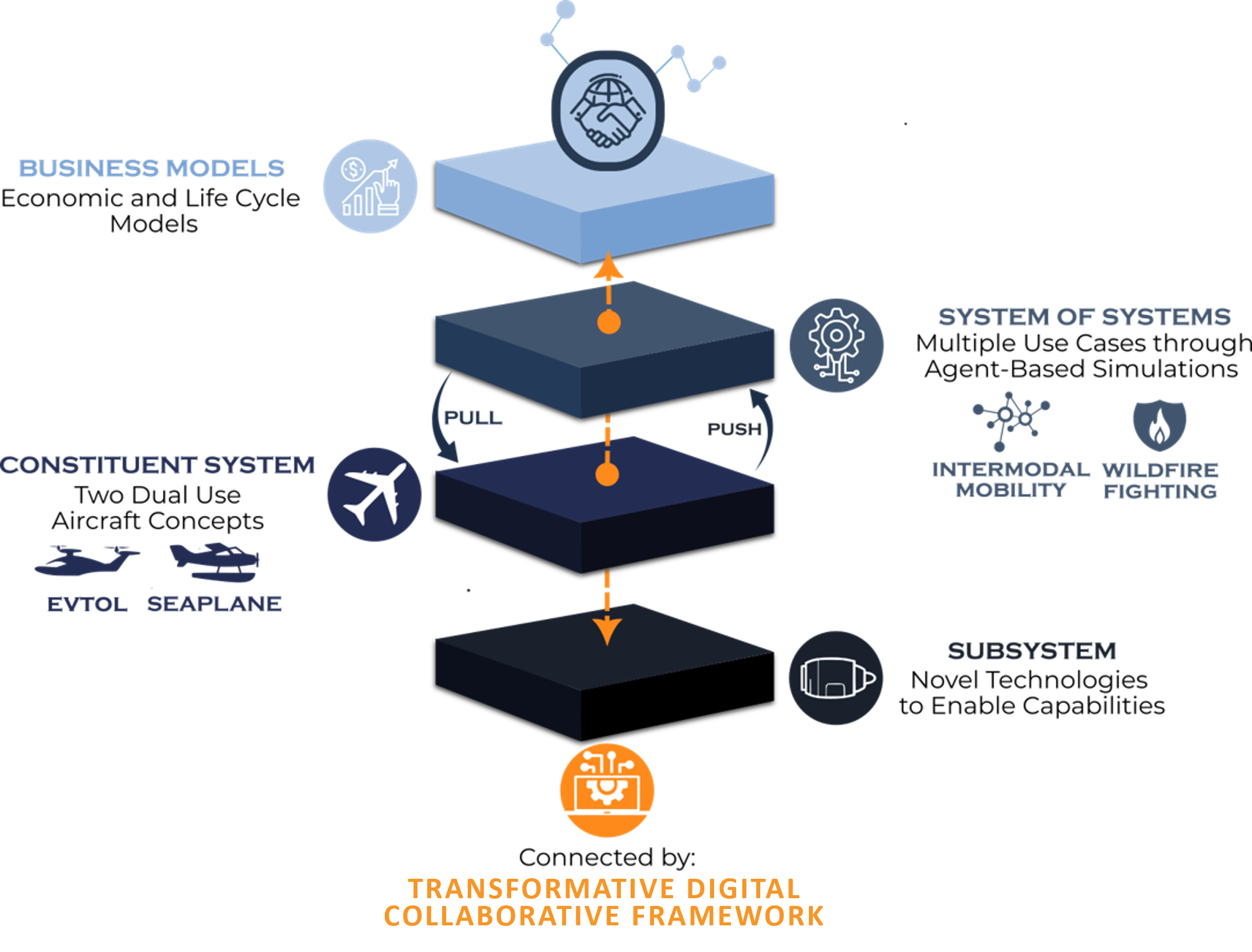}
\end{center}
\caption{COLOSSUS EU project overview~\cite{shiva2024colossus}.}
\label{fig:COLOSSUS}
\end{figure}

In the context of the European Union (EU)–funded COLOSSUS project\footnote{\url{https://colossus-sos-project.eu/project/}} (Collaborative System-of-systems exploration of aviation products, Services and business models), led by DLR, several advanced methods for analyzing and optimizing complex System-of-Systems (SoS) architectures are under investigation~\cite{shiva2024colossus}. As aviation integrates ever more sophisticated technologies and operational concepts, a SoS perspective in which multiple autonomous systems coordinate to achieve capabilities unattainable by any one system alone becomes essential. This paradigm opens the door to novel solutions for contemporary challenges, such as Advanced Air Mobility and aerial wildfire fighting. %, when dealing with multi-agent systems.
Designing such complex SoS increasingly relies on integrated modeling and optimization techniques that can handle heterogeneous interactions and architectural complexity. Agent-Based Modeling (ABM) offers a powerful paradigm for representing SoS: each constituent system is portrayed as an autonomous agent with its own dynamics and decision rules, and the emergent global behavior arises from agent interactions and negotiations~\cite{varenne2013nature}. This bottom-up approach captures critical SoS traits such as operational and managerial independence, emergent behavior, and adaptive evolution, which makes ABM especially suited for early-stage SoS evaluation and design exploration~\cite{baldwin2015simulation}.
In practice, to support early-stage evaluation of SoS configurations, Agent-Based Simulations (ABS) provide a flexible way to capture interactions among heterogeneous platforms. Building on these simulations, our work proposes a holistic System-of-Systems Engineering %(SoSE) 
methodology developed within COLOSSUS~\cite{bussemaker2024system}. The goal is to explore alternative SoS architectures and, from these, to derive Top-Level Aircraft Requirements (TLAR) for innovative aerial systems. Our method follows a top-down %(\emph{product-pull}) 
paradigm: scenarios and Concepts of Operations drive ABS, whose outcomes guide hierarchical optimization of SoS architectures, which in turn inform TLAR for new aircraft concepts~\cite{villas2024concept} and can therefore be used for optimization or risk assessment~\cite{zimdahl2024risk}.
The COLOSSUS methodology is applied to two complementary use cases:
\begin{itemize}
  \item \textbf{Advanced Air Mobility (ADAM)}: a bottom-up (\emph{product-push}) approach to develop sustainable 4D intermodal mobility business models and vehicle concepts.
  \item \textbf{Eco-friendly Vehicle for multiple operating Environments (EVE)}: a top-down (\emph{product-pull}) approach to assess novel aircraft in the context of aerial wildfire fighting.  
\end{itemize}
 More precisely, both ADAM and EVE integrate two new “eco-aircraft” configurations specifically tailored for both very short-range passenger transport and wildfire response: 
\begin{enumerate}
  \item \textbf{All-electric VTOL}: a multi-role vertical take-off and landing aircraft targeting 4D intermodal mobility and rapid wildfire intervention, with an Entry Into Service (EIS) by 2030.
  \item \textbf{Multi-role Seaplane}: a CS23-class aircraft optimized for both mobility and firefighting. Two development horizons are considered: short-term EIS in 2035 and long-term EIS in 2050, with configurations tuned for commonality, manufacturability, and operational requirements.
\end{enumerate}
An overview of the COLOSSUS project is presented in Figure~\ref{fig:COLOSSUS}, which illustrates the four hierarchical layers of its transformative digital collaboration framework. From the bottom up, (i) Subsystems (including propulsion units, sensors, and onboard systems) generate performance data that is passed to (ii) Constituent Systems, representing individual aerial platforms such as eVTOLs and seaplanes. These platforms are then embedded within (iii) a System-of-Systems architecture, enabling coordinated intermodal operations for applications such as wildfire response and sustainable mobility, through bidirectional "push–pull" data flows. At the top level, (iv) Business Models and economic analyzes capture stakeholder priorities, cost–benefit trade-offs, and policy scenarios, providing strategic guidance for design, integration, and deployment~\cite{prakasha2024colossus}.%\vspace{-0.1cm}
In this paper, we will focus only on the EVE test case that is introduced in more detail in Section~\ref{sec:context} and on aircraft design and optimization in such a wildfire-fighting context. 
The remainder of this paper is organized as follows. 
Section~\ref{sec:context} describes the EVE use case for which hierarchical optimization of SoS architectures is applied. Section~\ref{sec:smt} introduces surrogate modeling for expensive black-box functions with hierarchical design variables. Section~\ref{sec:optim} presents our multi-objective, constrained Bayesian optimization framework with mixed hierarchical decision variables. 
Section~\ref{sec:results} provides the optimization results obtained on EVE studies.
Finally, Section~\ref{sec:conclusion} concludes and outlines perspectives for further applications in intermodal mobility.

\section{The COLOSSUS EVE use case}
\label{sec:context}
\subsection{EVE  presentation}
\label{EVE}

\begin{figure}[H]
\begin{center}
\includegraphics[width=\linewidth]{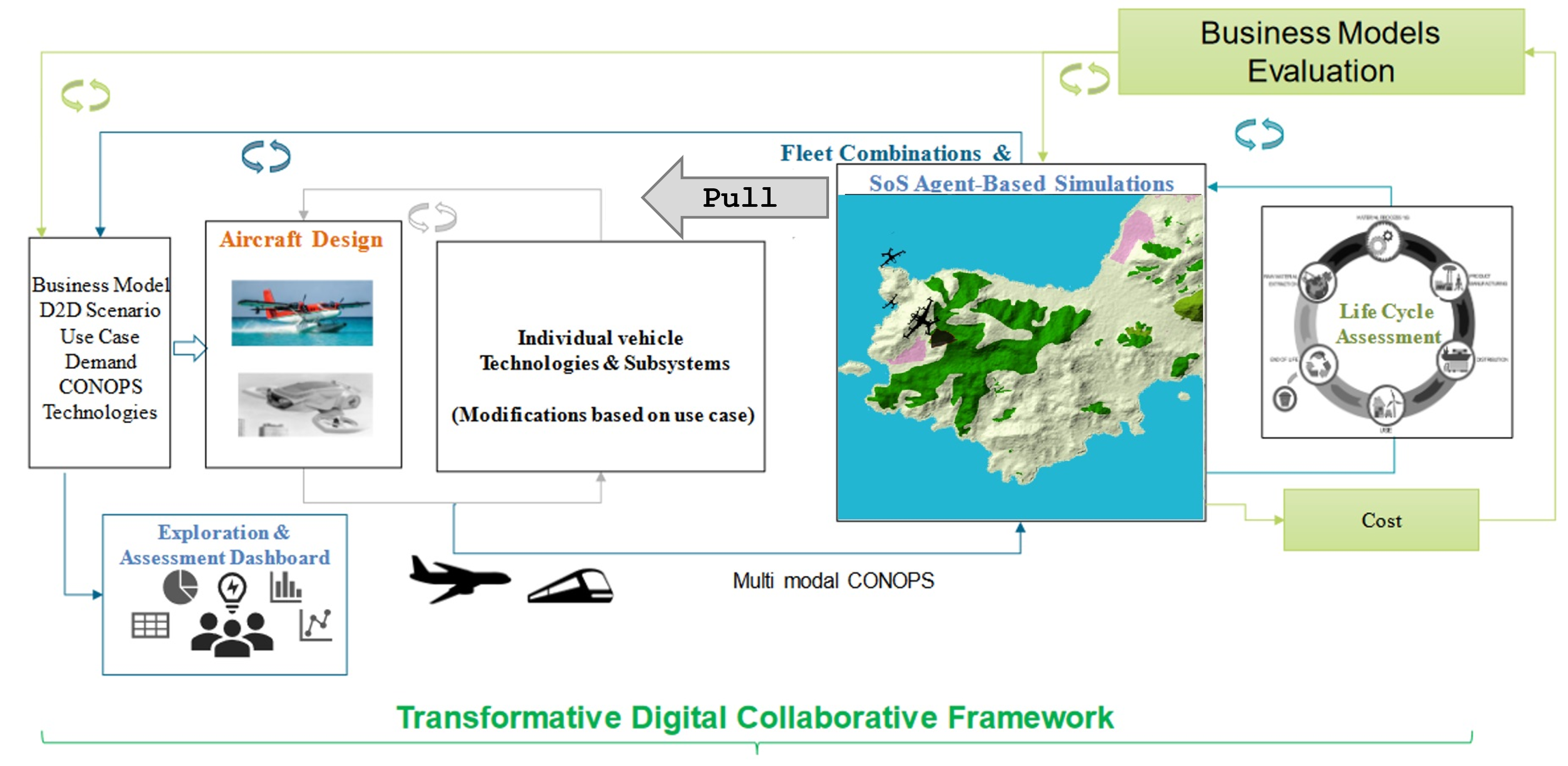}
\end{center}
\caption{The COLOSSUS pull paradigm based on EVE.}
\label{fig:eve}
\end{figure}

The EVE use case, whose methodology is illustrated in Figure~\ref{fig:eve}, utilizes a \emph{product-pull} strategy. Rather than designing an aircraft in isolation, this approach extracts the necessary Top-Level Aircraft Requirements (TLARs) directly from the overarching operational needs of the System-of-Systems (SoS). To achieve this, the method uses agent-based simulations of the entire SoS, capturing the whole wildfire-fighting fleet interacting within its dynamic environment. Within this holistic setup, the mission tactics—or Concepts of Operations (ConOps)—and the core vehicle design variables are optimized simultaneously. Consequently, the final aircraft specifications emerge naturally; rather than being arbitrarily pre-defined, the vehicle's requirements are directly driven by what the mission actually needs to succeed~\cite{naeem2024productpull}.  

%\vspace{-0.1cm}

The ConOps represent how aircraft coordinate, prioritize, and allocate resources and, as such, are essential to the overall effectiveness of the design method~\cite{maier1998architecting}. Each candidate architecture and  ConOps pairing is evaluated in an Agent-Based Simulation (ABS) that models agent behavior, fire propagation, and inter‐agent interactions. The agent-based simulation outputs feed value functions (\textit{e.g.}\ burned‐area, resource efficiency, environmental impact) to drive the optimization loop~\cite{bussemaker2024ADSG,deleeuw2022}. 
It is important to note that "optimization" in this context does not imply finding a singular, absolute optimum for the entire SoS; rather, it refers to exploring design solutions that offer better overall performance relative to specific criteria and stakeholder needs—solutions we characterize as "optimized" to align with traditional engineering terminology~\cite{prakasha2024colossus}. A multi-objective global optimizer then discovers a Pareto Front (PF) of ConOps that are the "optimal alternatives" design combinations. From this frontier, stakeholder priorities and constraints select a final solution, whose performance requirements become the TLARs for new aircraft design~\cite{villas2024concept}.   
In Agent-Based Modeling (ABM), where emerging sociotechnical phenomena are studied, simulations often model ill-defined and context-sensitive dynamics where controlled experimentation is infeasible~\cite{angione2022}.  ABS offers a bottom‐up means to emulate complex, adaptive systems ranging from disease spread to economic markets and urban growth by representing each entity as an autonomous agent whose local interactions give rise to global behavior~\cite{bremer2016}. A typical ABS comprises agents that follow either fixed rule sets or adapt through learning, an environment that defines the spatial or networked backdrop for agent encounters, and interaction protocols that dictate how agents perceive, decide, and influence one another~\cite{michel2018}. Depending on their design, ABS may rely on explicit rule‐driven logic, stochastic decision‐making, or hybrid mechanisms combining both deterministic and probabilistic elements~\cite{taillandier2019}. As the number of agents and the density of their interactions increase, the computational burden escalates rapidly, making detailed, large‐scale studies often impractical~\cite{macal2010tutorial}. Moreover, calibrating these models can require extensive parameter sweeps with potentially thousands of runs, to ensure fidelity, and slight variations in inputs can yield significantly divergent outcomes, complicating uncertainty quantification~\cite{schulze2017advances}. To address these challenges, surrogate modeling techniques are employed to construct inexpensive approximations of the ABS that preserve its essential dynamic patterns, thereby enabling efficient exploration, sensitivity analysis, and optimization without the full computational cost of the native simulation~\cite{forrester2008engineering, saves2025modeles}. 
An agent can, for example, correspond to a firefighting aircraft or a group of ground units. Consequently, ABS provides an effective means of modeling and simulating both SoS architectures and scenario dynamics within a unified, high-level abstraction~\cite{saves2025surrogate}.
ABS allows for the investigation of the components in an SoS Architecture, from the number and composition of constituent systems, the ConOps, to the degree of control over the constituent systems.

Representing a SoS through an ABM approach requires representing the key actors of the SoS, their ConOps, and, in particular, their interactions and individual processes. Each actor’s role and relationships with others must be explicitly captured in the ABM, just as they would be in the real SoS. In addition, the environment in which the SoS operates must be represented with sufficient fidelity to reflect both system–system and stakeholder–environment interactions~\cite{saves2024system}. 
Once defined, the ABM is executed through an ABS providing insights into the value functions associated with scenario objectives. Within the COLOSSUS project, these simulations are conducted using the System-of-Systems Inverse Design (SoSID) toolkit~\cite{kilkis2021python}, a Python-based ABS environment dedicated to the emulation of SoS that was specifically developed for use cases such as EVE.
%aerial wildfire suppression (EVE) and urban air passenger transportation (ADAM).
ABS captures bottom-up emergence effectively; its flexibility and opacity raise credibility and interpretability challenges, presenting barriers for decision support and policy alignment~\cite{blanco-volle2024}.
The challenges of credibility and interpretability in ABS are not unique to agent-based approaches; many stochastic black-box simulators face similar issues~\cite{palar2026interpretable}. To manage these, formal modeling, verification, and trade-space analysis are employed, not only to evaluate designs but also to shape possibilities in response to evolving objectives and constraints~\cite{chan2025goal}.  It is important to note the critical role of high‐fidelity ConOps modeling in ABS for capturing dynamic coordination in SoS, ensuring TLARs reflect both mission dynamics and emergent fleet behavior~\cite{bussemaker2024ADSG,saves2023smt}. 

To effectively model and optimize complex aerial operations, our proposed framework is structured around the continuous exchange of information across three distinct, nested levels of abstraction~\cite{naeem2024productpull}. At the highest tier, the fleet level represents the overarching System-of-Systems. It accounts for the complex interactions, dynamic coordination strategies, and emergent phenomena that arise when multiple autonomous agents operate collectively within a changing, real-world environment.  The intermediate tier, known as the constituent system level, models the individual aircraft, such as an eVTOL or a seaplane, acting as an autonomous agent within the fleet. This level maintains a critical, bidirectional coupling with the SoS level above it~\cite{villas2024concept}. The constituent systems feed essential aircraft data, such as performance metrics and physical geometry, up to the SoS simulation. In return, they receive operational feedback and updated %Top-Level Aircraft Requirements 
TLARs, including target flight times and ranges, which are dictated by the overall mission needs. 
Finally, the foundational tier is the subsystem level, which encompasses the specific technological and physical components housed within each aircraft. This includes hardware such as propulsion units, sensors, and onboard systems that generate the baseline performance data enabling the constituent system to function~\cite{akbari2025architecting}.
%
%Therefore, the simulation relies on a multi-level approach to account for the interactions and to propagate the information across the different levels.
%The coupling between the aircraft agent level and the SoS level is of highest priority, as SoS like EVE focus on this interaction. The coupling occurs in both directions: the SoS level relies on data from the aircraft, such as performance metrics, geometry, and other characteristics, to incorporate them into simulations. Conversely, the aircraft constituent systems agents also expect feedback from the SoS level, such as updates on TLARs, including average flight time and range.
Additionally, other data from the SoS level, such as the number of selected aircraft, are valuable to design an optimal system.

\subsection{EVE formalization}
\label{Formalization}
The SoS problem exhibits more complexity than the usual system optimization. For instance, it can be characterized by a multi-level aspect with strong coupling between the SoS level and System level (agent design), of intermediate dimension in terms of variables (up to 100), with mixed variables (including categorical and hierarchical choices), and with several objectives (linked to stakeholder key performance indicators) and constraints. In addition, the solvers involved (SoS simulation, for instance) will hardly provide derivative information and will be costly to operate. Furthermore, architectural choices as well as ConOps parameters are bound to be part of the optimization variables~\cite{bussemaker2024system}.
In addition to these general requirements, the EVE use case adds specific ones. For example, the optimization approach will focus on the exploitation of ConOps, where existing firefighting strategies can be represented and simulated. The approach relies on the selection of suppression tactics, which are predefined behaviors followed by the agents. These suppression tactics prioritize different firefighting strategies, such as targeting fire fronts with the highest spread rate, those closest to flammable vegetation, those near elevation slopes that could accelerate fire spread, or creating a fire block/ellipse around the fire. These suppression tactics can be considered as simplified configuration variables within an SoS architecture, resulting in a categorical choice for the optimizer. To increase the realism of the study, an additional level of complexity is introduced by incorporating alternative tactics that are triggered under specific conditions in the simulation (\textit{e.g.}, distance to fire, time limits). This introduces a hierarchical decision-making process based on the initial tactic selection and predefined swap criteria and thresholds~\cite{saves2024system}.
Therefore, the optimization approach must be capable of handling such a complex and constrained design space. This will enable the exploration of tactics trees that represent ConOps.

\section{Efficient surrogate models of hierarchical systems}
\label{sec:smt}

Designing and optimizing architectures in complex SoS contexts often relies on computationally demanding simulations, including ABS, where derivatives are not available, and each run can be prohibitively costly. Surrogate models address this challenge by acting as lightweight approximations of the original simulator, making it possible to explore the design space more efficiently and to support hierarchical optimization tasks such as system parameter tuning or aircraft design~\cite{llacay2025}. By learning from pre-computed simulation datasets, surrogates can approximate response surfaces at a fraction of the computational expense, thereby accelerating analysis and enabling scenario exploration that would otherwise remain infeasible under strict resource constraints~\cite{deleeuw2022}. Their use has already been shown to be effective in extracting response surface models from agent-based simulations, supporting both large-scale experimentation and rapid decision-making~\cite{fabiani2024}.

\subsection{Hierarchical surrogate}
In the context of hierarchical SoS design, the goal is to optimize value functions derived from ABS with respect to selected system parameters and aircraft design variables using a black-box model~\cite{Lambe2011,Lambe2012,Lambe2013}. This model is generally a computationally expensive-to-evaluate (where derivatives are not available) problem that could be encountered in industry. Therefore, it could be useful to use a surrogate model that reduces the computational cost while giving a good approximation of the black-box simulation.  The surrogate is often built from a small, expensive-to-evaluate dataset called the Design of Experiments (DoE) representing the set of known configurations~\cite{saves2025modeles}. 
Nevertheless, the process generally involves mixed continuous-categorical design variables as design variables cover both continuous and discrete domains to represent the wide range of decisions involved in configuring a complex fleet. Continuous variables can describe characteristics such as aerial agent endurance, maximum speed, or payload capacity that are parameters that can be tuned finely within physical and operational limits. Discrete variables, on the other hand, capture high-level configuration choices, such as the number of drones deployed or categorical choices such as the selection of firefighting tactics.  
Furthermore, hierarchical dependencies frequently emerge in problems involving variable dimensions or exclusive technological alternatives. For instance, high-level operational choices, such as the overall firefighting strategy, can dictate the required number and type of aircraft. Similarly, selecting an eVTOL platform architecture decrees battery-specific design parameters, whereas these variables would be inactive in a seaplane scenario. Because certain high-level decisions decree the presence or absence of subordinate parameters, the resulting design space is structured.

Surrogate models, also known as metamodels, are indispensable in engineering design optimization for approximating expensive black-box simulations while drastically cutting computational cost~\cite{Rasmussen,Martins2021}.  Therefore, in this work, we aim to construct an inexpensive surrogate model $ \hat{f}$ for a black-box simulation $f$ depending on $d$ variables. 
The function $f$ is typically expensive-to-evaluate simulations, with no exploitable derivative information.
Gaussian Processes (GP)~\cite{Rasmussen}, also called Kriging models, are known to be a good modeling strategy to define response surface models. 
Namely,
we will consider that our unknown black-box  function $f$ is a realization of an underlying GP of mean $\hat y$  and of standard deviation $s$, \textit{i.e.}.,
\begin{equation}
f \sim \hat{f}=\mathcal{G} \mathcal{P}\left(\hat y, s^2\right). \label{eq:GP:f}\end{equation}
For a general problem involving categorical or integer variables, we proposed a dedicated GP in~\cite{Mixed_Paul}, extended to high dimensions in~\cite{Mixed_Paul_PLS}, and to hierarchical variables in~\cite{saves2023smt}. 

In SoS, certain variables, termed \textit{meta-variables}, govern the presence or absence of other variables, effectively imposing a hierarchy or graph structure over the design space~\cite{saves2025hierarchical}. These structures are found in numerous engineering domains, including modular system design, configurable architectures, and reconfigurable vehicle platforms. They are especially of high interest for heterogeneous multi-agent systems such as the ones we are interested in, as we have different types of agents: eVTOLs and seaplanes.
Hierarchical surrogate modeling techniques exploit conditional or tree-structured input domains to create surrogate models as precise as possible without loss of information, and recent advances have extended GP surrogates to mixed continuous, integer, categorical, and hierarchical variables, jointly modeling meta-variables, decreed variables, and context-dependent activation functions~\cite{saves2023smt,halle2024graph,Architecture}.
Also, a unified surrogate modeling framework has been proposed and implemented in SMT 2.0, the Surrogate Modeling Toolbox~\cite{halle2024graph, saves2025hierarchical}. 
%SMT is an open-source modeling software that implements state-of-the-art techniques for constructing, evaluating, and comparing surrogate models across a wide range of applications, with an important focus on GP and on providing surrogate models derivatives\footnote{\url{https://smt.readthedocs.io/}}~\cite{saves2023smt, SMT2019}.
SMT is an open-source library for building and comparing advanced surrogate models, specializing in GP and derivative computation\footnote{\url{https://smt.readthedocs.io/}}~\cite{saves2023smt, SMT2019}.  

As aforementioned, we introduce meta-variables that determine whether other variables, called decreed variables, are active. In contrast, non-hierarchical variables are said to be neutral. The framework also supports {partially decreed variables}, which are conditionally activated depending on contextual logic, allowing for flexible modeling of subsystem configurations. 

\subsection{Wildfire fighting tactics example}

\begin{figure}[htb!]
\centering
\includegraphics[width=\linewidth]{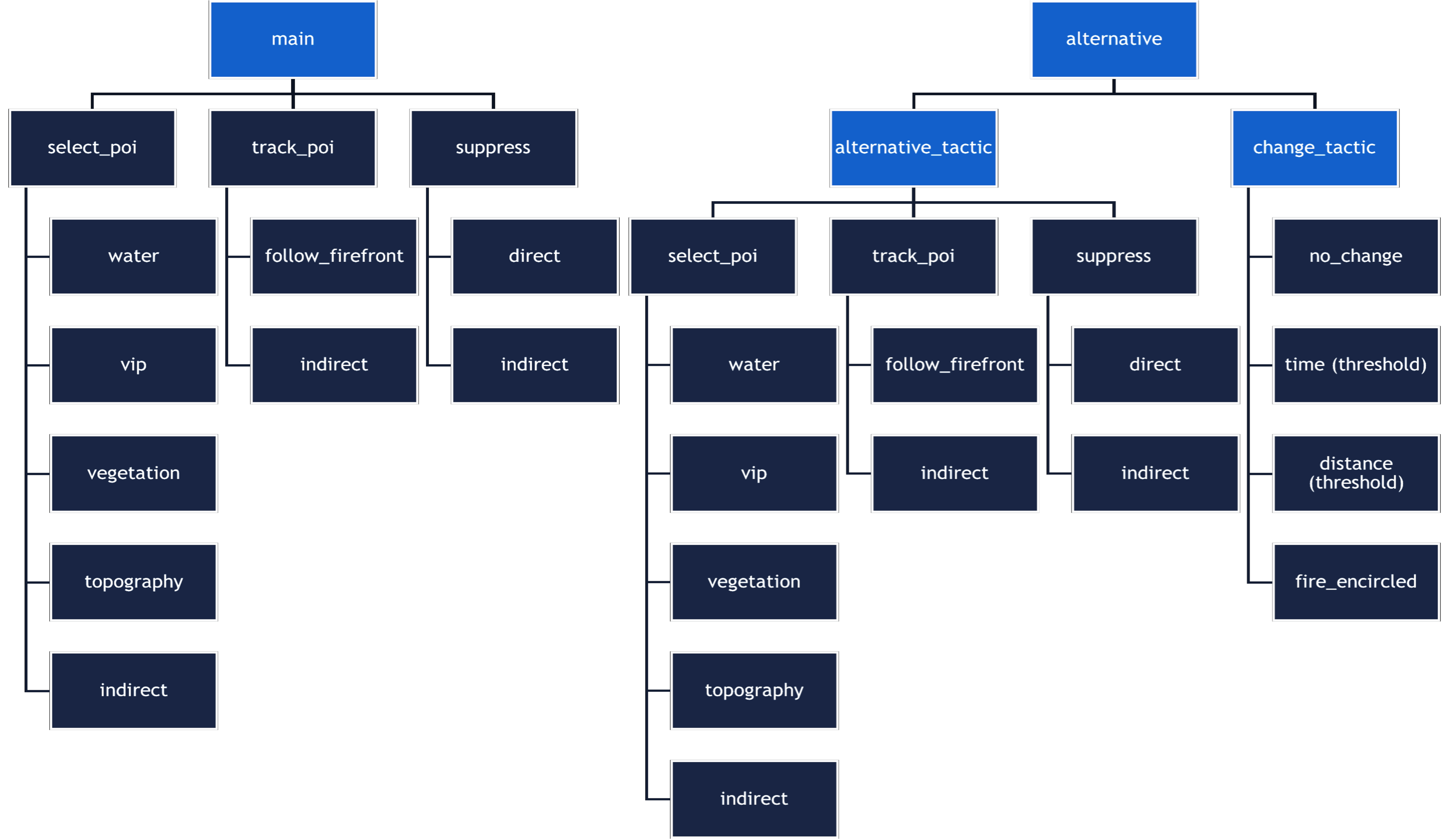}
\caption{Wildfire fighting with mixed hierarchical components.}
\label{fig:tree_hier}
\end{figure}

To demonstrate the application of our approach, we consider a wildfire-fighting ConOps centered around the selection and coordination of mission tactics. The scenario includes a set of hierarchical decisions spanning high-level mission objectives, operational contingencies, and asset configurations. For example, the root node \texttt{main} represents the primary firefighting tactic selected for the mission. 
More precisely, the tactics used are as follows: \emph{indirect attack}  applies a fire line logic, creating an ellipse around the center of the fire and attempting to surround the fire, \emph{water} simply chooses fire based on spread rate and closest water sources, \emph{vip} does the same as water but for pre-defined prioritizes areas, \emph{topography}   prioritizes fire suppression when elevation is more inclined to accelerate fire spread and \emph{vegetation}  does the same with vegetation instead.
The \texttt{alternative\_tactic} node encodes predefined contingency plans, specifying fallback options in case the primary approach becomes infeasible due to shifting fire behavior, resource availability, or safety concerns. The \texttt{change\_tactic} node governs the conditions and logic for switching from the primary to an alternative tactic, triggered by variables such as distance to fire front, time, or amount of specific area burnt.
Figure~\ref{fig:tree_hier} presents this tactical reasoning structure as a directed, tree-like hierarchy of system-level variables.
Each light blue node denotes a \emph{choice}, a high-level decision or functional block (\textit{e.g.} \texttt{main}, \texttt{alternative\_tactic}, \texttt{change\_tactic}), that “activates” one or more child (darker blue) variables or levels. Although this schematic does not yet distinguish between selecting a specific variable value and the underlying conditional activation logic, it provides a simple, intuitive way to enumerate and check on all of the variables influencing the problem.  Additionally, each edge in the tree indicates a dependency: when a parent choice is enabled, all of its connected child variables are conditionally activated.  
%Therefore, hereafter,  we will refine this representation by introducing explicit activation logic and value domains using our Design Space Graph formalism to fully capture both the selection of variable values and the hierarchical constraints between them.  

\begin{figure}[htb!]
\centering
\includegraphics[width=\linewidth,height=9cm]{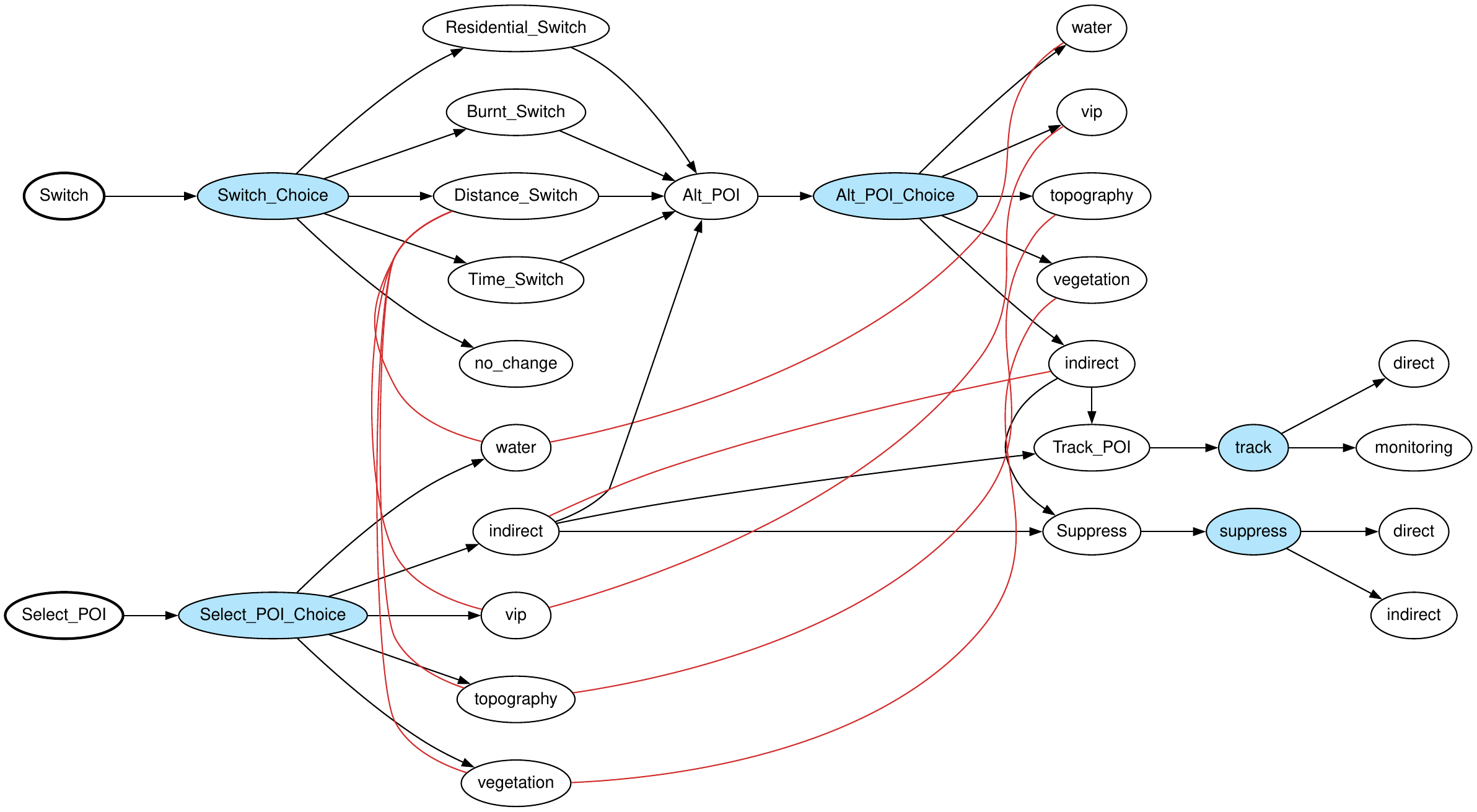}
\caption{Mixed hierarchical variables associated with the ConOps in the design space graph. Red lines are incompatibilities, arrows are decreed dependencies, and light blue nodes are selection choices from which exactly one output arrow is selected. }
\label{fig:adsg_example}
\end{figure}

Building on this intuitive tree representation, we next introduce a rigorous, graph-based formalism that both generalizes hierarchical and conditional domains and natively supports mixed-type variables. This formalism is automatically instantiated via the Architecture Design Space Graph (ADSG) software (adsg-core)~\cite{bussemaker2024ADSG}\footnote{\url{https://adsg-core.readthedocs.io}}. 
To illustrate these concepts on the wildfire-fighting agent system example, Figure~\ref{fig:adsg_example} shows the mixed and hierarchical variables Design Space Graph (DSG) hierarchy.
%in the Architecture Design Space Graph (ADSG) for our wildfire-fighting problem ConOps.
Mathematically, we use a general DSG, denoted $\mathcal{G}$, with both the TLAR for the aircraft design and several other continuous variables (such as timing for switching from one strategy to another). %More details about the problem and its variables are given further in the next sections.
This figure illustrates how multiple meta-variables govern the activation of decreed variables and their value choices among levels.  For instance, the \texttt{Switch\_Choice} meta-variable can invoke one of several switch tactics (\textit{e.g.}\ \texttt{Residential\_Switch}, \texttt{Burnt\_Switch}, \texttt{Distance\_Switch}, \texttt{Time\_Switch}, or \texttt{no\_change}) because multiple arrows originate from it, and the node is in light blue because it is a choice node.  Downstream, the \texttt{Alt\_POI\_Choice} meta-variable is conditionally inactive when \texttt{no\_change} is selected, but in turn governs alternative points of interest such as \texttt{water}, \texttt{vip}, or \texttt{topography}.  Red arcs indicate additional incompatibility dependencies, \textit{e.g.}\ if \texttt{Distance\_Switch} is chosen, a subset of POI types is unavailable, letting \texttt{indirect} as the only possible choice.  %
Here, the red lines encompass the incompatibility constraints that prohibit the alternative choice from being the same as the original one if activated by the switch choice (\textit{i.e.} when the threshold is attained and \texttt{no\_change} is not chosen).
This example highlights how the DSG formalism captures rich, cross-hierarchy interactions and variable-sized decision spaces in a single unified graph. 

Based on the hierarchical structure encoded in the DSG, we define a general way to compute distances between hierarchical configurations. This is essential for building effective surrogate models, particularly Gaussian Processes (GPs), in domains where conditional and hierarchical dependencies lead to heterogeneous input spaces.  To address this, we adopt the graph-structured distance introduced in~\cite{halle2024graph}. The method operates over an \emph{extended domain} in which all potential variables are considered, regardless of whether they are active in a given design instance. The distance metric accounts for three possible scenarios for each variable:
\begin{enumerate}
    \item \textbf{Both variables active}: a standard distance is computed (\textit{e.g.}, Euclidean for continuous variables, Hamming for categorical ones).
    \item \textbf{Both variables inactive}: the contribution is zero, as both designs agree that the variable is irrelevant.
    \item \textbf{One variable active, one inactive}: a fixed penalty is applied to reflect a structural mismatch in the design hierarchy.
\end{enumerate}
The overall distance between two design configurations is then computed by aggregating the contributions of each variable (\textit{e.g.}, via an $\ell_p$ norm). This distance respects the conditional activation patterns encoded by the DSG and is well-defined (that is, it satisfies metric properties such as symmetry and the triangle inequality). Then, the graph-structured distance serves as the foundation for defining kernels over structured design spaces. In SMT~2.0, we leverage this distance within standard kernels, such as the squared exponential or Matérn kernel, by substituting the graph-structured distance $d(x, x' |  \theta)$ knowing the vector of hyperparameters $\theta$ into the kernel function:
\begin{equation}
    k(x, x') = \exp\left( - {d(x, x' | \theta)^2} \right), \quad \forall x \in \mathbb{R}^d, x' \in \mathbb{R}^d, \theta\in \mathbb{R}^d 
    \label{eq:kernel}
\end{equation}
ensuring that the resulting kernel remains positive-definite.
By incorporating variable activeness directly into the distance computation, the surrogate model naturally reflects structural differences between design configurations. For example, two configurations that activate different subsystems will be far apart in the input space, while small changes within active variables lead to proportionally smaller distances. This modeling approach is seamlessly supported in SMT~2.0, which automatically manages the hierarchical activation structure via the ADSG software. Practitioners can construct GPs over such spaces without manually handling conditional variables, thanks to the integrated graph-based kernel support\footnote{\url{github.com/SMTorg/smt-design-space-ext}}.
More precisely, there are various ways to ensure that the kernel $k$ is well-defined and symmetric positive definite when dealing with graph-structure distances. In SMT 2.0, two variants, the Arc-kernel and the Alg-kernel, are implemented. Other kernels, such as the imputation kernel, are also discussed in~\cite{saves2023smt}.
These hierarchical GP models enable accurate surrogate modeling in highly structured domains, such as aircraft design, modular architecture optimization, or other complex SoS or multi-agent problems. 

In summary, the combination of graph-structured distance metrics and hierarchical GP kernels
yields a versatile surrogate modeling framework capable of accurately capturing both conditional activations and inter‐variable dependencies. This unified approach has been shown to significantly improve prediction quality and optimization convergence in mixed‐variable, tree‐structured domains.  
This modeling strategy aligns well with the COLOSSUS project’s objective to optimize SoS architectures, where each subcomponent (\textit{e.g.}, airframe, propulsion, mission subsystem) may activate different variables depending on high‐level design choices~\cite{saves2025hierarchical}. The SMT 2.0 toolbox, in conjunction with agent‐based simulations and architecture generators, thus forms a foundation for exploring large, structured design spaces with computational efficiency. In the next section, we extend this surrogate modeling framework to support \textit{multi-objective} and \textit{constrained} optimization, incorporating mixed-variable decision hierarchies using Bayesian optimization strategies.

\section{Multi-objective constrained Bayesian optimization with mixed hierarchical decisions}
\label{sec:optim}

\begin{figure}[H]
\centering
\includegraphics[width=0.8\textwidth]{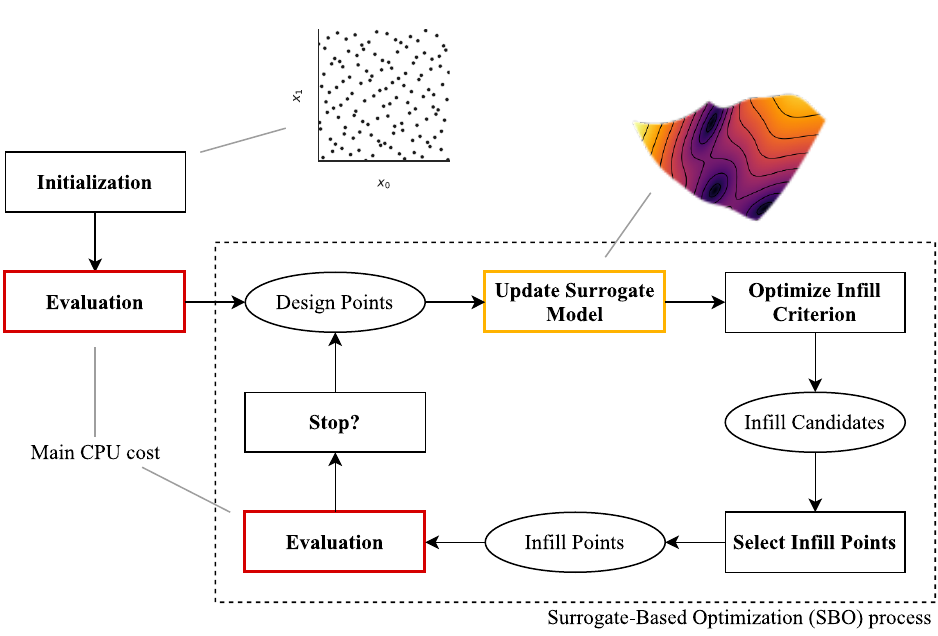}
\caption{The Bayesian optimization framework~\cite[Figure 2]{bussemaker2024system}.}\label{fig:sbo}
\end{figure}

Once the GP model is defined, we can use it for infill search and for optimizing an expensive black-box function through adaptive Bayesian Optimization (BO). In other words, we want to optimize a function by evaluating it with as few costly evaluations as possible. This process naturally balances the classic exploration–exploitation trade-off: early in the search, BO favors exploration by sampling regions of high uncertainty to improve the global surrogate, whereas later iterations exploit areas with predicted high performance to refine the Pareto front.  
BO relies on GP models of the objective and constraint functions to efficiently determine the most interesting design point to evaluate next. Once the infill point has been evaluated using the expensive evaluation functions, the surrogate model is reconstructed, and the process starts over until some termination criterion has been reached. Critically, BO does not seek a single optimum but, in the multi-objective setting, we seek a Pareto front of non-dominated trade-offs, where no objective can be improved without worsening another.  
Figure~\ref{fig:sbo} visualizes the working principle of BO, with, highlighted in red, the costly evaluations whose number is limited to a certain computational budget and in orange the GP surrogate models from SMT for objective and constraint functions. 

The implementation of this optimization process is based on a sequential enrichment approach, typically the Efficient Global Optimization algorithm~\cite{jones1998efficient} or Super EGO~\cite{Sasena02flexibility}, an evolution of EGO to handle constraints. BO is based on GP defined by Eq.~(\ref{eq:GP:f}), and here the idea is to use some adaptive mixture of GP-based models to tackle hierarchical problems. The mixture of experts (MOE)~\cite{dimitri:SM02011, Liem2015} is known to approximate complex functions with heterogeneous behavior by combining local surrogate models in a global one. 
The general framework based on SMT is called Super Efficient Global Optimization coupled with Mixture Of Experts (SEGOMOE)~\cite{bartoli:hal-02149236,bartoli2023multi,effectiveness,saves2022bayesian,grapin_constrained_2022}. 
In Section~\ref{subsection: Multi-objective Bayesian optimization}, we present previous works and our methodology for multi-objective Bayesian optimization, and, in Section~\ref{sec:segomoe}, we present our novel extension of SEGOMOE to handle mixed hierarchical constrained multi-objective optimization problems.

\subsection{Multi-objective Bayesian optimization}
\label{subsection: Multi-objective Bayesian optimization}

In various real-world problems, such as engineering design, resource allocation, and decision-making processes, there is often a need to simultaneously optimize multiple objectives that may compete or complement each other. This is where multi-objective optimization comes into play, offering a valuable framework for achieving a balance between adversarial goals and exploring trade-offs between them. By considering multiple objectives, the optimization provides a more comprehensive understanding of the problem landscape, enabling decision-makers to make more informed choices and enhance overall system performance. 
These optimization problems are difficult to solve, especially when the functions are expensive-to-evaluate black-boxes (complex and costly models) and have many local optima, which can be the case in the drone design context. Bayesian optimization is a powerful approach in this context.
Recently, work has been made to allow BO techniques to handle multi-objective optimization problems~\citep{grapin_constrained_2022}.
This section focuses on how to solve multi-objective optimization problems of the following form.

\begin{align}
    & \underset{x \in \Omega \subset \mathbb{R}^{d}}{\min} \quad \mathbf{f}(x):=[f_1(x),...f_n(x)]^T  \label{eq:pbmultiob} \\
    & \text{such that} \left \{\begin{array}{lr}
        \mathbf{g_{ieq}}(x) \geq 0   \\
        \mathbf{g_{eq}}(x)=0  \\
    \end{array} \right. \nonumber
\end{align}
where $d$ is the number of dimensions, $\mathbf{f}$ gathers the $n$ multiple objective functions $f_1$ to $f_n$, and the $m$ constraints are separated between $\mathbf{g_{eq}}$ representing equality constraints and $\mathbf{g_{ieq}}$ for inequality constraints.
Once the initial Design of Experiments is given in the design space, % in the relaxed space, 
GP-based surrogate models are built for the objective and constraint functions.  Each costly function $f_i(\x)$ is approximated by a GP characterized by its mean $\hyi: \mathbb{R}^{d} \to \mathbb{R}$ and its standard deviation ${s_i}(\x): \mathbb{R}^{d} \to \mathbb{R}$
    $$\hat{f}_i(\x)\sim \mathcal{N}(\hyi,\hsi) \quad i=1,\ldots, n$$
Regarding the multi-objective approach, we assume that the $n$ components of $\mathbf{f}$ are independent. This allows us to define $\hat{\mathbf{f}}$ as the surrogate model associated with each component, represented as:
$$\hat{\mathbf{f}}(\mathbf{x}) \sim \mathcal{N}(\hat{\mathbf{y}}(\mathbf{x}), \Sigma(\mathbf{x}))$$
In this equation, $\hat{\mathbf{y}}(\mathbf{x}): \mathbb{R}^{d} \to \mathbb{R}^n$ represents the GP prediction vector, given by $[\hat{y}_1(\mathbf{x}), \ldots, \hat{y}_n(\mathbf{x})]$. The matrix $\Sigma(\mathbf{x})$ is a diagonal matrix, with its diagonal elements provided by $s_i$, for all $i = 1, \ldots, n$.
Also, $\hat{g}_j(\mathbf{x}), \leq j \leq m $ represents the mean prediction of the $j^{\text{th}}$ GP constraint surrogate model.
The initial problem, as defined by Eq.~\eqref{eq:pbmultiob}, is substituted with an infill problem, which is formulated as follows:
\begin{equation}\label{eq:opt_acquisition_problem}
\left\lbrace
\begin{array}{l}
 \displaystyle \max_{{x}\in \mathbb{R}^{d}} \quad \alpha^{\mbox{reg}}_{\mathbf{f}}(\mathbf{x}) \\
\mbox{subject to} \\
        \mathbf{\hat{g}_{ieq}}(x) \geq 0   \\
        \mathbf{\hat{g}_{eq}}(x)=0  \\
\end{array}\right.
\end{equation}
In this formulation, $\alpha^{\mbox{reg}}_{\mathbf{f}}(\mathbf{x})$ is the regularized acquisition function as referenced in~\cite{saves2022bayesian, bartoli2023multi} relative to different hypervolume Improvement based criteria such as EHVI~\cite{zitzler2003performance,emmerich2006single}), PI~\cite{jones2001taxonomy}), or MPI~\cite{rahat2017alternative}).
To address the problem outlined in~\eqnref{eq:opt_acquisition_problem}, various optimization algorithms can be employed. These algorithms should be capable of handling nonlinear constraints and can be based on either derivative-free optimizers, such as COBYLA (Constrained Optimization BY Linear Approximation~\cite{cobyla94}), or gradient-based methods, such as SLSQP (Sequential Least Squares Programming~\cite{kraft1988software}) or SNOPT (Sparse Nonlinear Optimizer~\cite{gill2005snopt}). Moreover, a multi-start strategy can be utilized in conjunction with these algorithms.
SEGOMOE further embeds constraint surrogates to ensure that infeasible designs are automatically penalized, so the algorithm prioritizes candidate points that are both feasible and Pareto‐efficient.  To do that, our approach for the constraints relies on the mean criterion as in~\eqnref{eq:opt_acquisition_problem} because only the mean prediction of the constraints surrogate models is used, but less restrictive alternatives are also available in SEGOMOE~\cite{SEGO-UTB, SEGO-UTB-Bombardier}.  
This adaptive optimization process is iterated until the total computational budget is exhausted. Upon completion, the feasible points within the final database represent the identified Pareto optimal points.

\subsection{SEGOMOE with mixed hierarchical decisions}
\label{sec:segomoe}

Now, we generalize the methodology presented above for handling multiple objectives and constraints, but with more general inputs featuring mixed hierarchical variables. 
As aforementioned, this new approach is based on SEGOMOE, making it particularly suitable for design problems involving mixed continuous, integer, and categorical variables, including hierarchical and conditional dependencies between variables.
We generate the initial DoE using a Latin Hypercube Sampling (LHS) that we adapted for mixed-variable spaces~\cite{saves2023smt}. While we have also developed adaptations of other sampling methods, such as Sobol' sequences for mixed hierarchical variables~\cite{bussemaker2024system},   we consistently employ LHS in this study. This choice is due to its low-discrepancy properties, which, although slightly less optimal for variance approximation and uncertainty quantification compared to Sobol' sequences, offer a practical balance between efficiency and simplicity~\cite{l2000variance}.
After that, the optimization process relies on the GP surrogate models that are adapted to take into account both the mixed and the hierarchical structure within the correlation kernel~\cite{BaBuDiHwMaMoLaLeSa2023, halle2024graph}. Mixed correlation kernels consist of representing the correlations for the categorical variables as matrices, while the integer variables can be continuously relaxed~\cite{BaDiMoLeSa2023}. Many matrix modeling choices with different complexities can be made depending on the model precision and computational time desired. The continuous variables can be reduced in number by Kriging with Partial Least Squares (KPLS) to reduce the GP training time, which can be important because the model is trained at every optimization step~\cite{Bouhlel2016improving}. The categorical variables hyperparameter matrices can also be approximated thanks to Partial Least Squares regression as sub-matrices of lower dimensionality for high-dimensional Bayesian optimization~\cite{BaDiMoLeSa2023}.   During experiments, the chosen number of principal components within KPLS models does not exceed 4 or 5. 
Crucially, hierarchical relationships among variables are encoded as a directed graph and are captured by a graph-aware distance metric, which is incorporated into the correlation kernel.  The resulting hierarchical kernel (\textit{e.g.}\ the Alg‐kernel) naturally respects active/inactive dependencies between meta-variables and their child variables~\cite{saves2025hierarchical}. Algorithm~\ref{algo:segomoe_mixed_hier} illustrates how, in each iteration, SEGOMOE constructs mixed‐hierarchical GP surrogates for every objective and constraint, formulates and maximizes a regularized multi‐objective acquisition function under surrogate constraints to select the next sample in the original design space.

\smallskip
\begin{algorithm}[H]
\SetAlgoLined
{\textbf{Inputs:}}  Initial DoE $\mathscr{D}_0$ and set $t=0$;   
% \newline

\While{the stopping criterion is not satisfied}{
\vspace{.2cm}
\begin{adjustwidth}{0pt}{40pt}
\begin{enumerate}
    \item  Relax the integer variables continuously and assemble the categorical correlation matrices, compute the distance between the points knowing the graph-structured design space $\mathcal{G}$ \;
    \item Build the GP model for the $n$ objective function $f_i(\x)$ and the $m$ constraints $g_j(\x)$ based on the hierarchical kernels of~\eqnref{eq:kernel} over the design space $\mathcal{G}$; % and compute an estimation of the search space $\Omega_f$;
    \item Build the acquisition function $\alpha^{\mbox{reg}}_{\f}(\x)$ and let $\Omega_f$ be the admissible search space defined by the constraint surrogates $\hat{g}_j(\x)$ and a certain criterion ;
    \item Maximize the acquisition function within the feasible domain  $\Omega_f$: $$\x_{t}:=  \underset{\x \in \Omega_f}{\arg \max}  \  \alpha^{\mbox{reg}}_{\f}(\x)$$
    \item Add $\x_{t}$, $f_1(\x_{t}), \ldots f_m(\x_{t})$, and $ g_1(\x_{t}), \ldots g_m(\x_{t})$ to the DoE  $\mathscr{D}_{t+1}$. Increment $t$;
\end{enumerate}
\end{adjustwidth}
\smallskip
}
{\textbf{Outputs:}}  The PF database of non dominated points %and the predicted PF; 
\caption{SEGOMOE for constrained multi-objective and mixed-integer problems.}
 \label{algo:segomoe_mixed_hier}
\end{algorithm}

\color{black}

\section{Application to wildfire fighting - EVE}
\label{sec:results}

\subsection{Wildfire optimization with hierarchical variables }

\begin{figure}[H]
\centering
\includegraphics[width=\linewidth]{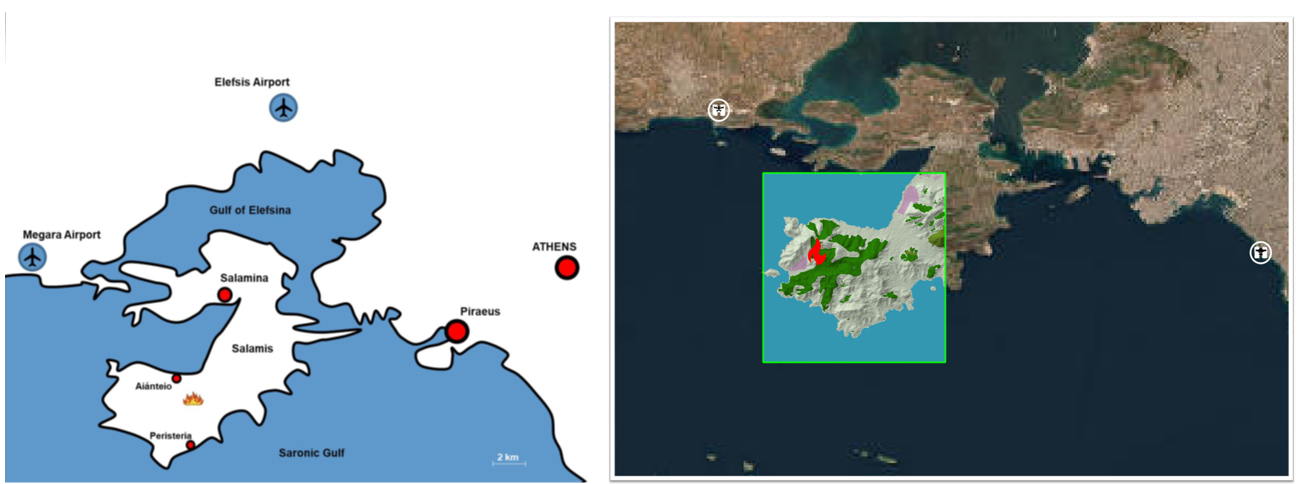}
\caption{EVE scenario - Salamis Island.}\label{fig:eve_sal}
\end{figure}

Within the frame of the project, two firefighting scenarios are investigated %in optimization studies. 
The first one focuses on a wildfire occurring on the island of Salamina, west of Athens, during a hot, sunny summer day with light wind conditions. The second one deals with a wildfire in the Pyrénées National Park, close to Luz-Saint-Sauveur, still in the summer period. Having two completely different scenarios allows for better coverage of the needs of different stakeholders. In fact, one scenario is set in a “Landlocked” environment, and the other in a “Seaside” environment. These two environments emphasize different aspects of the wildfire-fighting problem. The former may have more man-made support infrastructure close by, while natural water bodies may be scarce. The latter may lack close access to support infrastructure, but water from the sea is accessible, depending on weather conditions.
Figure~\ref{fig:eve_sal} illustrates the Salamis scenario location, along with an example of its recreation within SoSID. 
%In the framing of the project, only 
Two types of agents are considered in both scenarios: an eVTOL fleet and a seaplane fleet. Each type of agent is initially stationed at its own airbase. Modifiable scenario parameters include the location of the ignition center, the ignition start day and time, and the response time required to activate the fleet. 

 The simulation within SoSID uses a grid-based fire model based on cellular automata~\cite{rui2018forest}, which advances the fire front at each time step according to the surrounding environmental conditions, including terrain combustibility, elevation slopes, and weather data. Aircraft are represented as agents whose technical specifications and operational procedures, such as suppression tactics and operational constraints, are defined based on user input. SoSID currently supports energy, fuel, and hybrid aircraft architectures.
 Regarding optimization studies, only Salamina scenario investigations are presented in the paper as the approach is similar between both scenarios. 
 %Regarding optimization settings,
 In addition to the variables of the \textbf{suppression tactic} of wildfires presented in Section~\ref{sec:smt}, several other variables are included in the optimization problem: 
\begin{itemize}
  \item \textbf{Fleet size:} the number of suppressing aircraft within the operation is a variable of the problem for both eVTOL and seaplane fleets located on two different airbases.
   \item \textbf{Agent TLARs:} In the pull paradigm, the SoS-level optimization directly incorporates TLARs as problem variables, fully integrating them into the agent performance predictions. Once the optimal constellation is identified, these derived TLARs are extracted to guide the subsequent design phases for the seaplanes and eVTOLs.

  \end{itemize}
The objective functions are directly derived from the Key Performance Indicators (KPIs) of the various stakeholders, which collectively quantify the overall success of the scenario~\cite{zimdahl2024risk}.
In this study, three objective functions are considered:
\begin{itemize}
 % \item Fire emissions that evaluate the CO2 emissions of the fire according to the total amount of burnt area
  \item \textbf{Fire cost} that computes the overall cost of the fire damage. Each cell of the map is associated with a type of terrain (vegetation, forest, urban, ...), each having its specific cost value. 
 % \item Network Fuel that takes into account the operational aspect of the firefighting through the fuel consumed by the agents (here only the seaplanes, as the EVTOL is fully electric)
 \item \textbf{Operational cost} that takes into account the operational aspect of the firefighting through the overall mission cost of all the involved agents. As the seaplane operating cost is roughly three times higher than the eVTOL, the fleet composition will have a direct impact on this output.
 \item   \textbf{Operational emission}s that are directly related to the amount of fuel consumed by the agents (here, only the seaplanes will have an effect, as the eVTOL concept is fully electric).
  \end{itemize}
To limit the computational time of each evaluation, a fixed square area that covers a large portion of the map is defined as the external boundary for both scenarios. The fire is not expected to spread beyond this limit; if it does, the mission is considered a failure. This condition is treated as a constraint in the optimization process.

\begin{figure}[htb!]
\centering
\includegraphics[width=\linewidth,height=9cm]{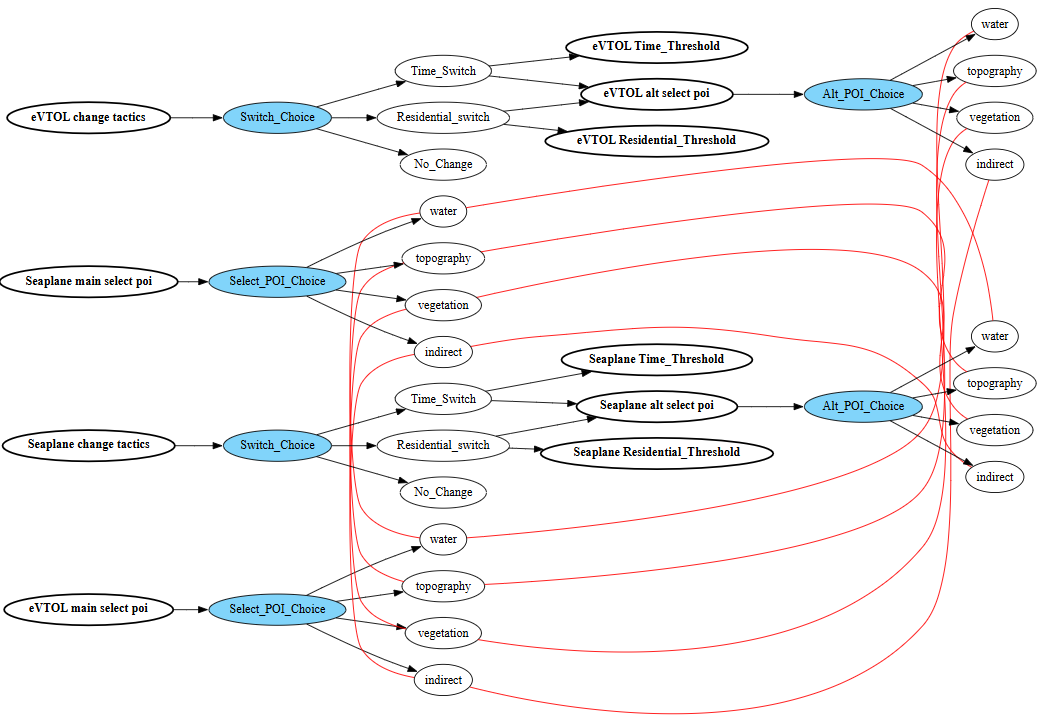}
\caption{Mixed hierarchical variables associated with the ConOps for the optimization problem of Table~\ref{tab:pb_hier} represented in the design space graph. }
\label{fig:adsg_2}
\end{figure}

 Table~\ref{tab:pb_hier} recaps the variables, objective, and constraint functions of the optimization problem, including the hierarchical variables. 
 In addition, several constraints are added to this design space regarding the tactics:
 \begin{itemize}
     \item The alternative strategy should always be different from the main one for each type of agent.
     \item The switch threshold being a decreed variable, only the one corresponding to the change tactics switch is activated for each type of agent.   
     \item Last, the main strategy (main select\_poi) should always be different between the seaplane and eVTOL agents.
 \end{itemize}
Figure~\ref{fig:adsg_2} presents a more detailed view of the mixed and hierarchical variables related to the \textbf{suppression tactic} handling. Compared to Figure~\ref{fig:adsg_example}, some simplifications have been made with the track\_poi and suppress choices fixed and the removal of \emph{vip} choice for both \texttt{main\_tactics}, \texttt{alternative\_tactic} (as no \emph{vip} area is currently used in the scenario). Nevertheless, the optimization problem will have to consider these \textbf{suppression tactic} for both agent types with some additional incompatibility constraints.

\begin{tableplain}[!htb]
    \caption{EVE Hierarchical problem.}   % <-- visible top caption
    \label{tab:pb_hier}
  \centering
      \small
    \begin{adjustbox}{max width=\linewidth}
    \begin{tabular}{llcrrl}
        \toprule
        & Function/Variable  & Type & Quantity & Range & Role \\
        \midrule
        \textbf{Minimize} & Fire cost & Continuous & 1 & & \\
       % & Fire cost & Continuous & 1 &  & \\
      %  & Casualties & Continuous & 1 & & \\
        & Operation cost & Continuous & 1 &  & \\
        & Operation emissions & Continuous & 1 &  & \\
        \addlinespace
        & \multicolumn{2}{l}{\textbf{Total objectives}} & \textbf{3} &  & \\
        \midrule
        \textbf{With respect to}  
         & EVTOL Range & Continuous &1&[50,200]  &  Neutral\\
          & EVTOL Speed & Continuous &1& [28,56]  &  Neutral\\
           & EVTOL Payload & Continuous &1& [340,580]  &  Neutral\\          
        & Seaplane Range & Continuous & 1 &[190,390]  &   Neutral\\
        & Seaplane Speed & Continuous & 1 & [70, 145]  &   Neutral\\
        & Seaplane Payload & Continuous & 1 & [1235, 2225]  &   Neutral\\
        & EVTOL fleet size & Ordinal & 1 & [4,6,8,10,12] & Neutral \\
        & Seaplane fleet size & Ordinal & 1 & [6,8,10] & Neutral \\
        & EVTOL main select\_poi & Categorical & 4 levels & & Meta \\
        & EVTOL change\_tactics & Categorical & 3 levels & & Meta \\
        & EVTOL alt select\_poi & Categorical & 4 levels & & Partially decreed\\

        & EVTOL distance switch threshold & Continuous & 1 & [1,5] & Decreed \\
        & EVTOL runtime switch threshold & Ordinal & 1 & [50,100,150] & Decreed \\
        & Seaplane  main select\_poi & Categorical & 4 levels & & meta \\
        & Seaplane  change tactics & Categorical & 3 levels & & Meta \\
        & Seaplane  alt select\_poi & Categorical & 4 levels & & Partially decreed\\

        & Seaplane distance switch threshold & Continuous & 1 & [1,4] & Decreed \\
        & Seaplane runtime switch threshold & Ordinal & 1 & [100,200,300] & Decreed \\
        
        \addlinespace
        \cmidrule(lr){2-4}
        & \multicolumn{2}{l}{Total continuous variables} & 8 & & \\
        & \multicolumn{2}{l}{Total integer and ordinal variables} & 4 & & \\
        & \multicolumn{2}{l}{Total categorical variables} & 6 & & \\
      %  \cmidrule(lr){2-4}
      %  & \multicolumn{2}{l}{\textbf{Total relaxed variables}} & \textbf{28} & & \\
        \midrule
        \textbf{Subject to} & \multicolumn{2}{l}{Mission success} & 1 & & \\
        & \multicolumn{2}{l}{\textbf{Total constraints}} & \textbf{1} & & \\
        \bottomrule
    \end{tabular}
\end{adjustbox}
  \end{tableplain}
  
\subsection{Salamis scenario results}

\paragraph{Baseline results}
A baseline (or reference result) is required in order to assess the improvement of the COLOSSUS solution on a similar scenario using state-of-the-art wildfire fighting agents and tactics. In this study, since only aerial agents are studied (and implemented), the DHC-515 aircraft has been selected for the baseline. In addition, only a conventional water attack strategy is considered. Therefore, in the Salamis use case, the reference scenario involves 2 DHC 515 using water tactics with an initial response time of 4 hours.
\begin{figure}[!htb]
   \begin{subfigure}[b]{.49\linewidth}
      \centering
      \includegraphics[width=\textwidth, height=6cm]{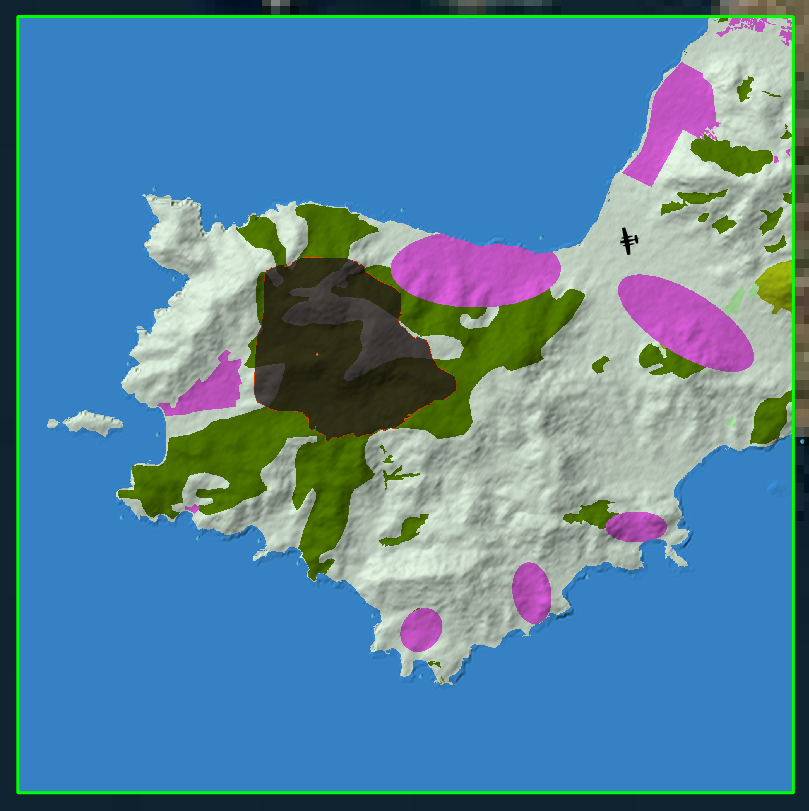}
      \caption{Fire propagation at response time (4 hours after fire ignition)}
      \label{fig:myfig15}
      \end{subfigure}
      \hspace{0.33cm}
      \begin{subfigure}[b]{.49\linewidth}
      \centering 
     {\includegraphics[width=\textwidth,height=6cm]{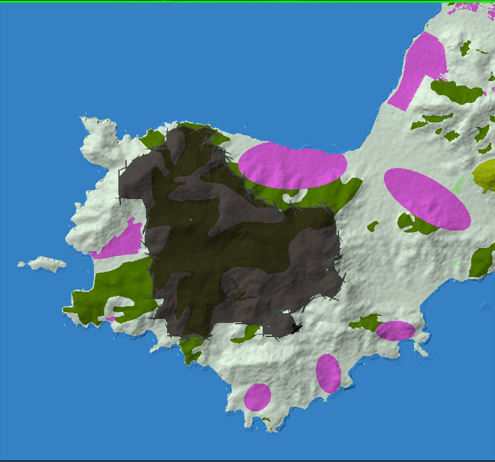}}
      \caption{Fire final state with baseline fleet (8 hours after response time)}
      \label{fig:myfig16}
   \end{subfigure}

   \caption{Baseline fleet solution on Salamis scenario}
   \label{fig:ref}
\end{figure}
 Figure~\ref{fig:ref} presents the results obtained with the baseline fleet of two DHC-515 aircraft. One can check that four hours after ignition (at response time), the fire extension is already quite advanced and getting close to the two cities area (identified by a purple color on the map), making this use case rather complex to solve. Nevertheless, the two DHC-515 fleet managed to stop the fire extension while protecting the cities after almost 8 hours of wildfire fighting. The performance of this baseline fleet, in terms of fire and operations, sets the reference values on which the COLOSSUS solutions will be benchmarked.

\paragraph{COLOSSUS results}

\begin{figure}[!htb]
   \centering

   \begin{subfigure}[b]{.49\linewidth}
      \centering
      \includegraphics[width=\textwidth,height=5cm]{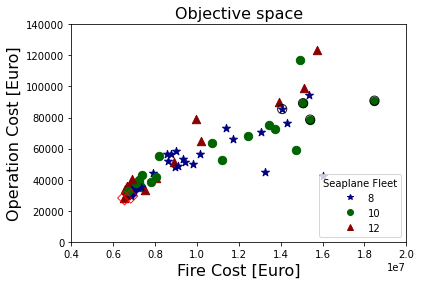}
      \caption{Operation Cost vs Fire Cost}
      \label{fig:myfig11}
   \end{subfigure}
   \hfill
   \begin{subfigure}[b]{.48\linewidth}
      \centering 
      \includegraphics[width=\textwidth,height=5cm]{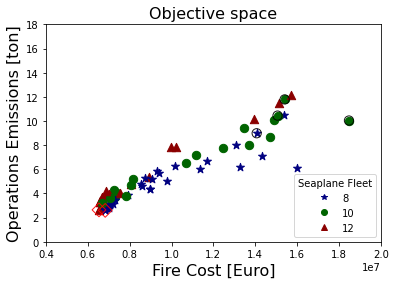}
      \caption{Operation Emissions vs Fire Cost}
      \label{fig:myfig12}
   \end{subfigure}

   \vspace{0.5em}

   \begin{subfigure}[b]{.51\linewidth}
      \centering 
      \includegraphics[width=\textwidth,height=5cm]{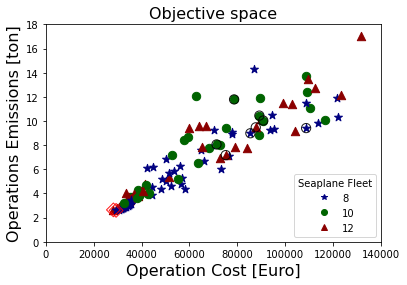}
      \caption{Operation Emissions vs Operation Cost}
      \label{fig:myfig13}
   \end{subfigure}

   \caption{Optimization results: initial DOE of 40 points (9 feasible identified by black circles) and 116 feasible enrichment points with 2 points on the Pareto Front identified by the red diamonds. The differences in shape and color depend on the number of seaplanes in the fleet (8, 10, or 12).}
   \label{fig:EVE_results_Sal}
\end{figure}

Regarding this preliminary optimization process, Figure~\ref{fig:EVE_results_Sal} illustrates the results obtained after an initial DOE of 40 points, followed by 170 BO iterations for a total of 210 expensive blackbox evaluations. 

\begin{figure}[!htb]
   \centering

   \begin{subfigure}[b]{.49\linewidth}
      \centering
      \includegraphics[width=\textwidth]{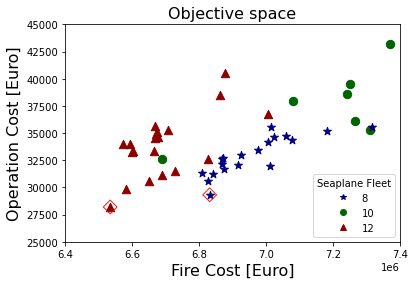}
      \caption{Operation Cost vs Fire Cost (zoom)}
      \label{fig:myfig21}
   \end{subfigure}
   \hfill
   \begin{subfigure}[b]{.48\linewidth}
      \centering 
      \includegraphics[width=\textwidth]{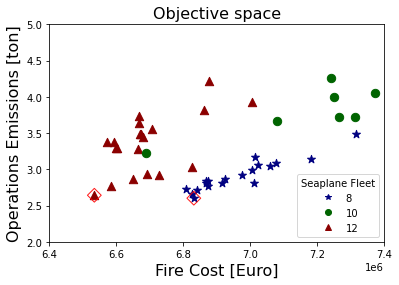}
      \caption{Operation Emissions vs Fire Cost (zoom)}
      \label{fig:myfig22}
   \end{subfigure}

   \vspace{0.5em}

   \begin{subfigure}[b]{.51\linewidth}
      \centering 
      \includegraphics[width=\textwidth]{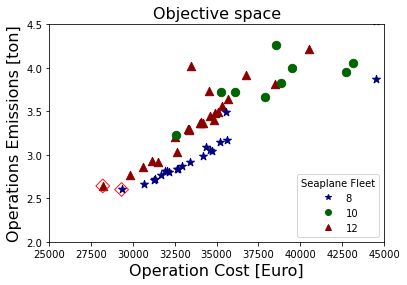}
      \caption{Operation Emissions vs Operation Cost (zoom)}
      \label{fig:myfig23}
   \end{subfigure}

   \caption{Optimization results: initial DOE of 40 points (9 feasible) and 101 feasible enrichment points with 5 points on the Pareto Front identified by the red diamonds (zoom).}
   \label{fig:EVE_results_Sal_zoom}
\end{figure}
Several insights can be drawn from these figures. First, among the initial DOE points, only a few (9) appear in the figure, highlighting the fact that most of them led to mission failure, which means that the combination of fleet and tactics was not able to stop the fire within the boundaries of the map, confirming the complexity of the firefighting. In addition,  these feasible DOE points are located mainly on the right-hand side of the figures; therefore, far from the current Pareto Front (PF). Therefore, the optimizer managed to successfully advance towards the lower values of all objective functions. Another interesting feature is the diversity of solutions explored by the optimizer with all three possible seaplane fleets (8, 10, or 12) populating the enrichment database. Eventually, the current Pareto Front, with 2 candidates, appears quite concentrated, and Figure~\ref{fig:EVE_results_Sal_zoom}  presents a zoomed version of the same dataset closer to the Pareto Front.
Two main clusters of solutions are identified, one based on a 12-seaplane fleet and the other on an 8-seaplane fleet.  As expected, the cluster with the lowest operation emissions corresponds to configurations using an 8-seaplane fleet, whereas the other cluster involving a 12-seaplane fleet has the highest firefighting efficiency. One can note that the 10-seaplane solutions (in dark green circle) are hardly found in the vicinity of the Pareto Front, with only one combination among the 12-seaplane cluster. 
 Table~\ref{tab:optimization_results} recaps the results obtained by the two PF configurations compared to the baseline fleet. Both COLOSSUS fleets enable a significant reduction in the burnt area and, consequently, the overall cost of the fire. Due to the fleet size being greatly increased compared to the baseline, the effective mission time is also divided by a factor of 3. Nevertheless, the operation emissions remain in favor of the baseline fleet, which contains only two agents. 
  
\begin{figure}[!htb]
   \begin{subfigure}[b]{.5\linewidth}
      \centering
      \includegraphics[width=\textwidth]{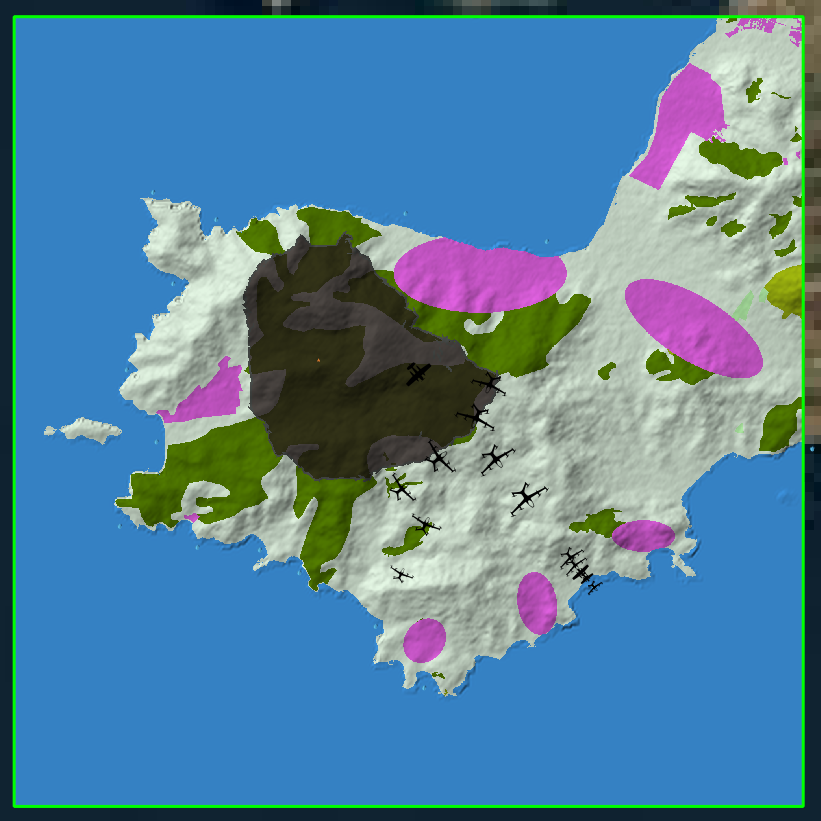}
      \caption{Fire final state with 8-seaplane fleet and associated tactics combination }
      \label{fig:myfig15}
      \end{subfigure}
      \begin{subfigure}[b]{.5\linewidth}
      \centering 
     {\includegraphics[width=\textwidth]{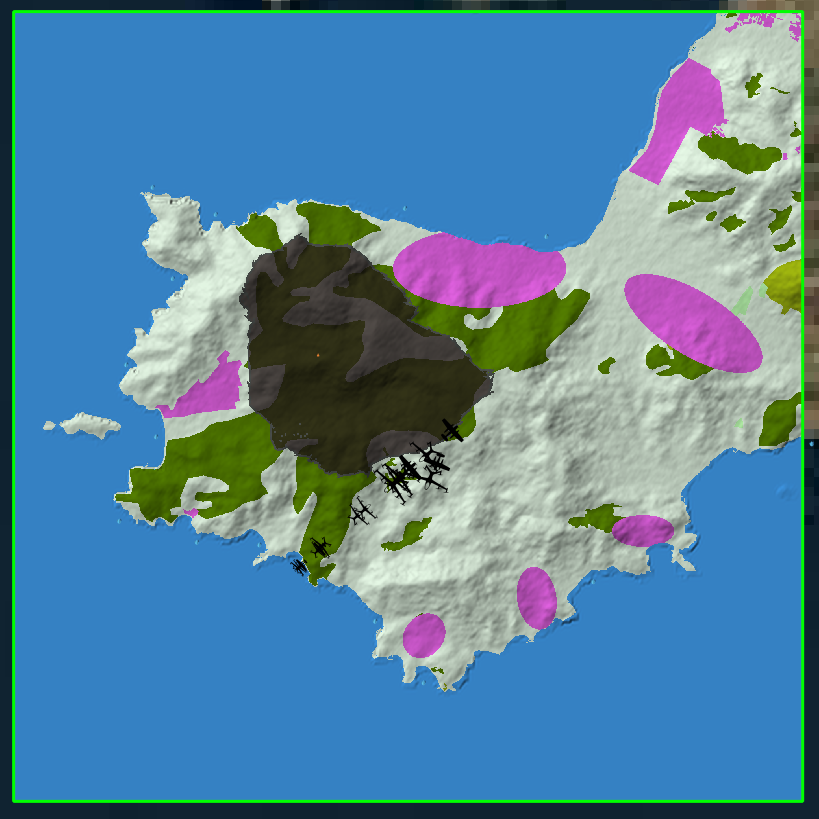}}
      \caption{Fire final state with 12-seaplane fleet and associated tactics combination }
      \label{fig:myfig16}
   \end{subfigure}

   \caption{Optimal fleet solutions on Salamis scenario}
   \label{fig:best}
\end{figure}

 Figure~\ref{fig:best} provides a visualization of the firefighting results obtained with the best-performing configuration using the 8-seaplane fleet (respectively, a 12-seaplane fleet). Compared to the baseline, the fire is extinguished before it reaches a significant size, with a small advantage to the 12 - seaplane fleet in terms of burnt area. 

\begin{table}[htbp]
\centering
\caption{Comparison of the baseline and optimized fleet configurations.}
\label{tab:optimization_results}
\begin{tabular}{lrrrrr}
\toprule
Configuration &
Total Fire Cost  &
CO$_2$ Emissions  &
Burnt Area  &
Effective Mission Time 
 \\
 &
 [M€] &
 [t] &
 [km$^2$] &
 [h] \\

\midrule
Baseline         & 11.53 & 1.63 & 11.79 & 7.68  \\
8 Seaplanes  &  6.83 & 2.60 &  6.67 &  2.  \\
12 Seaplanes &  6.53 & 2.64 &  6.37 &  1.6\\
\bottomrule
\end{tabular}
\end{table}

\begin{figure}[htbp]
    \centering

    \begin{subfigure}{0.48\textwidth}
        \centering
        \includegraphics[width=\linewidth]{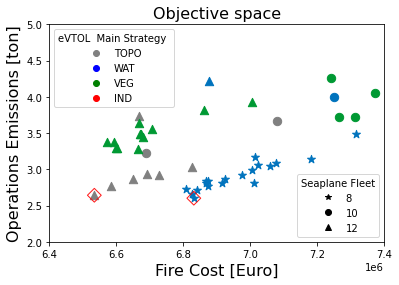}
        \caption{eVTOL main tactics}
        \label{fig:img1}
    \end{subfigure}
    \hfill
    \begin{subfigure}{0.48\textwidth}
        \centering
        \includegraphics[width=\linewidth]{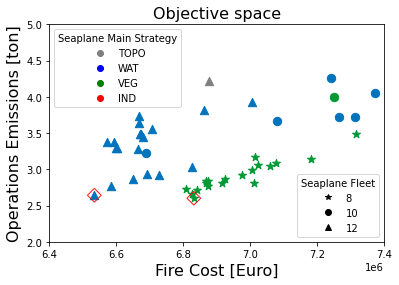}
        \caption{Seaplane main tactics}
        \label{fig:img2}
    \end{subfigure}

    \vspace{0.3cm}

    \begin{subfigure}{0.48\textwidth}
        \centering
        \includegraphics[width=\linewidth]{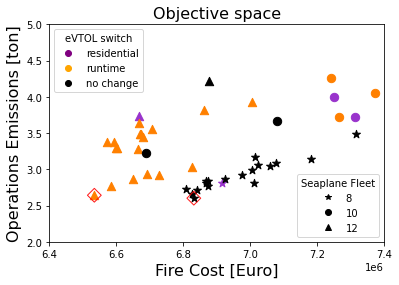}
        \caption{eVTOL switch}
        \label{fig:img3}
    \end{subfigure}
    \hfill
    \begin{subfigure}{0.48\textwidth}
        \centering
        \includegraphics[width=\linewidth]{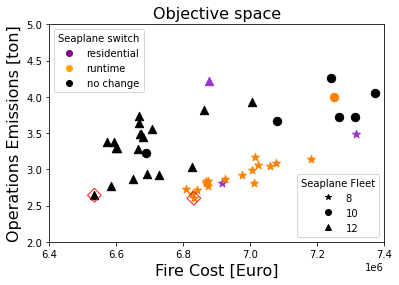}
        \caption{Seaplane switch}
        \label{fig:img4}
    \end{subfigure}

    \vspace{0.3cm}

    \begin{subfigure}{0.48\textwidth}
        \centering
        \includegraphics[width=\linewidth]{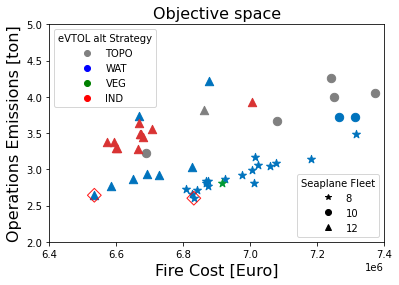}
        \caption{eVTOL alternative tactics}
        \label{fig:img5}
    \end{subfigure}
    \hfill
    \begin{subfigure}{0.48\textwidth}
        \centering
        \includegraphics[width=\linewidth]{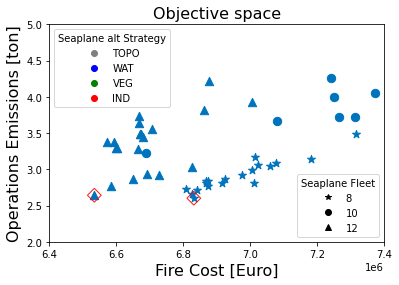}
        \caption{Seaplane alternative tactics}
        \label{fig:img6}
    \end{subfigure}

    \caption{Comparison of the six configurations.}
    \label{fig:comparison}
\end{figure}

%%%%%%%%%%%%%%%
 Figure~\ref{fig:comparison} represents, for the configurations close to the PF, the combination of \texttt{main\_tactics}, \texttt{alternative\_tactic}, and \texttt{change\_tactic} criteria for both agent types for the Fire Cost vs. Operations Emissions objective functions. The colors correspond to the \texttt{main\_tactics}, \texttt{alternative\_tactic}, and \texttt{change\_tactic}, and the mark shape to the number of seaplanes in the fleet (8, 10, or 12)
 Regarding the \texttt{main\_tactics}, the 8-seaplane eVTOL fleet relies on \emph{water}, and for the 12-seaplane eVTOL fleet, two clusters can be identified with \emph{topography} and \emph{vegetation}. Although all three possible \texttt{change\_tactic} choices are used, the  12-seaplane eVTOL fleet mainly selects \emph{runtime}, whereas almost all 8-seaplane eVTOL fleet use the \emph{no\_change} option, meaning that the \texttt{alternative\_tactic} will not be activated. The \texttt{alternative\_tactic} choice therefore exhibits less variety among all the fleets as the 12-seaplane eVTOL fleet clusters either select \emph{water} or \emph{indirect}, and the 8-seaplane eVTOL fleet kept the \emph{water} one. Only one constellation does not follow the pattern with \emph{residential} switch to \emph{vegetation}, showing that the optimization exploration still proposes an alternative solution.
 Regarding seaplane tactics choices, each fleet size relies on a different \texttt{main\_tactics}, namely \emph{water} for the 12-seaplane fleet and \emph{vegetation} for the 8-seaplane one. Most interesting is the \texttt{change\_tactic} choice and its consequence: the 12-seaplane fleet selects the \emph{no\_change} one; they keep using the \emph{water} tactics for all the missions. On the other side, the 8-seaplane fleet uses \emph{runtime} and all switch to \emph{water}, resulting in all the seaplanes (in the PF area) eventually using a similar tactic for the end of the mission.
 
\begin{figure}[htbp]
    \centering

    \begin{subfigure}{0.48\textwidth}
        \centering
        \includegraphics[width=\linewidth]{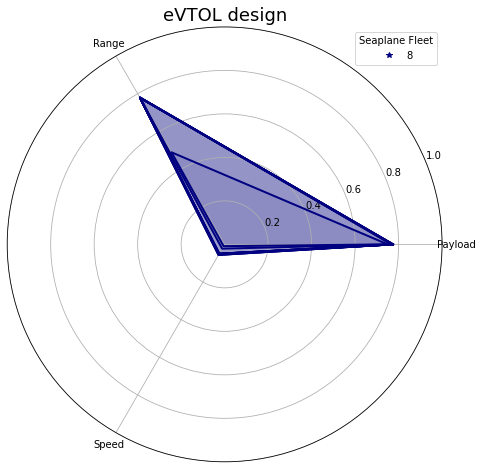}
        \caption{eVTOL design }
        \label{fig:img1}
    \end{subfigure}
    \hfill
    \begin{subfigure}{0.48\textwidth}
        \centering
        \includegraphics[width=\linewidth]{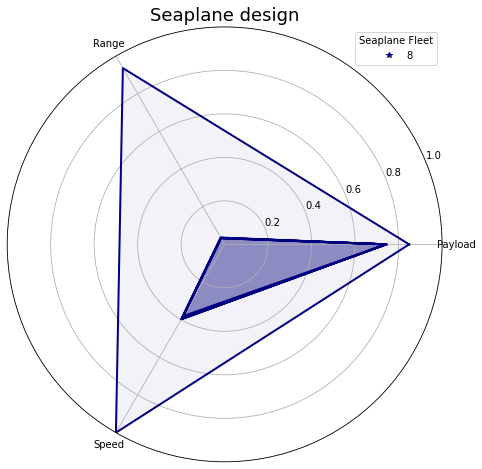}
        \caption{Seaplane  design}
        \label{fig:img2}
    \end{subfigure}

    \vspace{0.3cm}

    \begin{subfigure}{0.48\textwidth}
        \centering
        \includegraphics[width=\linewidth]{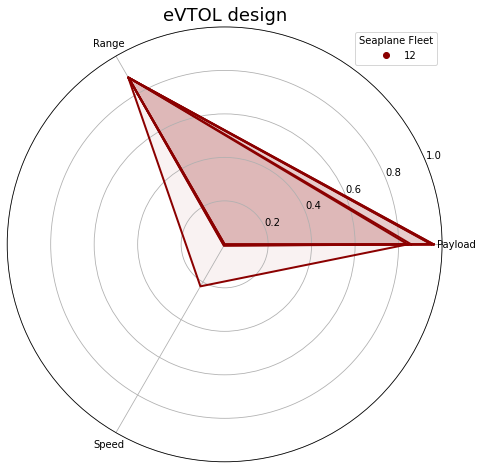}
        \caption{eVTOL  design}
        \label{fig:img3}
    \end{subfigure}
    \hfill
    \begin{subfigure}{0.48\textwidth}
        \centering
        \includegraphics[width=\linewidth]{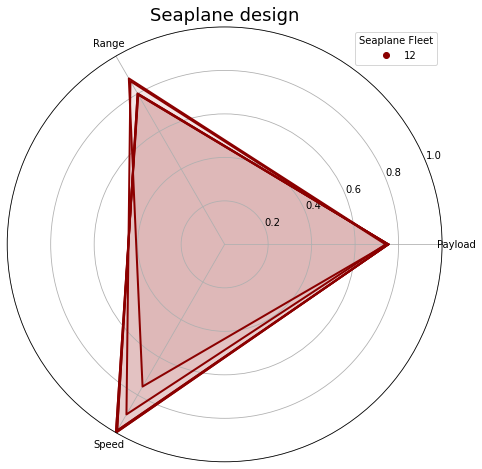}
        \caption{Seaplane  design}
        \label{fig:img4}
    \end{subfigure}

    \caption{Comparison of the six configurations.}
    \label{fig:design}
\end{figure}

 Last, Figure~\ref{fig:design} provides the TLAR information (in terms of payload, speed, and range) for the best-behaving agents for the 8-seaplane and 12-seaplane fleets. The selection criteria were to have a fire cost value below 7 million euros and an operation emissions value below 3 tons. Regarding the 12-seaplane fleet, 6 combinations of eVTOL and seaplanes are selected; for the 8-seaplane fleet, 11 combinations are considered. 
 For all selected solutions, the eVTOL fleet size is the biggest possible, with 12 agents. First, regarding eVTOL TLARs, the design is rather similar for all the selected combinations with high payload capacity and high range. The speed on the other side is close to the minimal value of the variable boundaries. 
 More interesting is the TLAR analysis of the seaplanes. Here, the design is more differentiated between the 8 and 12 -seaplane fleet. First 12-seaplane designs tend to have high speed, high range, and rather high payload, whereas almost all 8-seaplane fleet designs have low speed, minimal range, and medium payload. Only one design looks similar to the 12-seaplane fleet ones, but it corresponds to the solution where the \texttt{change\_tactic} criterion is \emph{vegetation} and not \emph{runtime} like the others. The combination of a bigger fleet and higher payload, range, and speed might explain the better performance of the 12-seaplane fleet compared to the 8-seaplane ones. 
 
To conclude this use case, the BO approach successfully demonstrated its capabilities to identify feasible and improved combinations of fleet size, TLARs, and tactics for a set of two different agent types. Results exhibited clusters of solutions, with different TLARs and tactics in the PF area, based on the number of seaplane agents in the overall fleet. For each set of clusters, best solutions are characterized by small variations of TLARs and \texttt{change\_tactic} criteria threshold values, but among those solutions, some alternative TLARs / tactics are also exhibiting good performance, highlighting the exploration capability of the BO algorithm.
Nevertheless, around the PF, the algorithm tends to mainly select  \emph{water} criteria for \texttt{main\_tactics} and \texttt{alternative\_tactic}. This can be explained by the fact that the fire is located on a peninsula with the sea surrounding the area, and also by the fact that, at response time,  the fire is close to both cities' area, preventing a choice like \emph{indirect} to be efficient in terms of fire cost. On the other hand, eVTOLs seem to have a broader choice of efficient tactics combination that they are mostly acting as a support role in this scenario (with no inland water sources, they can not be as efficient as seaplanes in terms of rotation time).

%%%%%

%\textcolor{red}{The complete optimization problem, including hierarchical variables, will be applied and analyzed in the final version of the paper.}
 %The two 12-seaplane configurations on the Pareto front are similar in terms of eVTOL fleet (10 eVTOL) and in terms of tactics combination and switch conditions. Small differences appeared in switch threshold values and agents' TLAR.  The same observations hold for the two 8-seaplane configurations of the PF (with 12 eVTOL). 
 %Comparing both solutions, the tactics combination is also different, with the 12-seaplane fleet starting with an indirect attack for seaplanes (and vegetation for eVTOL), whereas the 8-seaplane fleet relies on the vegetation as the main tactic for seaplanes and water for eVTOL. 
%\subsection{Pyrénées scenario results}

\color{black}

\section{Conclusion}
\label{sec:conclusion}

To conclude, this paper introduced a hierarchical Bayesian optimization framework for the design and optimization of complex aircraft-based multi-agent SoS. By combining Agent-Based Modeling to represent autonomous system behaviors with Agent-Based Simulations to evaluate scenario-level performance, the framework enables the exploration of large, heterogeneous, and structurally diverse design spaces with limited computational budgets. The use of Gaussian process–based hierarchical surrogate models allows efficient handling of mixed-variable, conditional, and hierarchical design spaces, addressing challenges commonly encountered in industrial-scale SoS optimization problems.

The framework was demonstrated through a wildfire-fighting ConOps use case within the EU-funded COLOSSUS project, coordinating heterogeneous aerial platforms to achieve mission objectives. Results show that the hierarchical surrogate approach improves both search efficiency and robustness compared to conventional surrogate-based methods, enabling the derivation of %Top-LevAel Aircraft Requirements 
TLAR for innovative aerial systems.

Overall, this work contributes to a scalable and transferable methodology for SoS architecting, bridging conceptual ABM representations, high-fidelity ABS evaluations, and surrogate-based optimization. Beyond wildfire fighting, the approach applies to a wide range of aviation, sustainable mobility, and resilience-focused system design challenges.

Future work will first focus on considering more than one scenario in the process, allowing different tactics to be combined for subgroups of agents. In a second step, it appears necessary to extend the approach to robust optimization in order to consider the uncertainties in the SoS simulation. Increasing the complexity of the simulation should also be considered with the extension of the framework to dynamic SoS environments, incorporating real-time adaptation and integrating multi-fidelity simulations to further reduce computational cost. %Additionally, expanding the methodology to include business models, safety constraints, and environmental considerations will strengthen its relevance for early-stage strategic decision-making in aircraft management and beyond.

\section*{Acknowledgements}
The research presented in this paper has been performed within the framework of the COLOSSUS project (Collaborative System-of-Systems Exploration of Aviation Products, Services and Business Models) and has received funding from the European Union Horizon Programme under grant agreement n${^\circ}$101097120.  This work is part of the MIMICO research project funded by the Agence Nationale de la Recherche (ANR) n$^o$ ANR-24-CE23-0380.  The authors are grateful to the partners of the COLOSSUS Project for their contribution and feedback.

\bibliography{output}

\end{document}